\documentclass[letterpaper]{article} % DO NOT CHANGE THIS
\usepackage{aaai2026}  % DO NOT CHANGE THIS
\usepackage{times}  % DO NOT CHANGE THIS
\usepackage{helvet}  % DO NOT CHANGE THIS
\usepackage{courier}  % DO NOT CHANGE THIS
\usepackage[hyphens]{url}  % DO NOT CHANGE THIS
\usepackage{graphicx} % DO NOT CHANGE THIS
\usepackage{natbib}  % DO NOT CHANGE THIS AND DO NOT ADD ANY OPTIONS TO IT
\usepackage{caption} % DO NOT CHANGE THIS AND DO NOT ADD ANY OPTIONS TO IT
\usepackage{algorithm}
\usepackage{algorithmic}
\usepackage{amsmath}
\usepackage{amssymb}
\usepackage{amsthm}
\usepackage{mathrsfs}
\usepackage{amsfonts}
\usepackage{bm}
\usepackage{soul}
\usepackage{xcolor}
\usepackage{booktabs} % 这行是关键！
\usepackage{graphicx} % 如果你使用了 \resizebox
\usepackage{float}    % 如果你用了 [H] 选项
\usepackage{colortbl}
\usepackage{pifont}
\usepackage{subcaption}
\usepackage{newfloat}
\usepackage{listings}
\newtheorem{theorem}{Theorem}

\newtheorem{definition}{Definition}

\DeclareCaptionStyle{ruled}{labelfont=normalfont,labelsep=colon,strut=off} % DO NOT CHANGE THIS
\floatstyle{ruled}
\newfloat{listing}{tb}{lst}{}
\floatname{listing}{Listing}
\title{HealSplit: Towards Self-Healing through Adversarial Distillation \\in Split Federated Learning}
\author{
	Yuhan Xie\textsuperscript{\rm 1,\rm 2},
	Chen Lyu\textsuperscript{\rm 1,\rm 2}\thanks{Chen Lyu is the corresponding author.}
}

\affiliations{
	\textsuperscript{\rm 1}Shanghai University of Finance and Economics\\
	\textsuperscript{\rm 2}MoE Key Laboratory of Interdisciplinary Research of Computation and Economics,\\ Shanghai University of Finance and Economics\\
	yhtse@stu.sufe.edu.cn, lyu.chen@sufe.edu.cn
}

\usepackage{bibentry}
\begin{document}

\maketitle

\begin{abstract}
Split Federated Learning (SFL) is an emerging paradigm for privacy-preserving distributed learning. However, it remains vulnerable to sophisticated data poisoning attacks targeting local features, labels, smashed data, and model weights. Existing defenses, primarily adapted from traditional Federated Learning (FL), are less effective under SFL due to limited access to complete model updates. This paper presents HealSplit, the first unified defense framework tailored for SFL, offering end-to-end detection and recovery against five sophisticated types of poisoning attacks. HealSplit comprises three key components: (1)  a topology-aware detection module that constructs graphs over smashed data to identify poisoned samples via topological anomaly scoring (TAS);   
(2) a generative recovery pipeline that synthesizes semantically consistent substitutes for detected anomalies, validated by a consistency validation student; and 
(3) an adversarial multi-teacher distillation framework trains the student using semantic supervision from a Vanilla Teacher and anomaly-aware signals from an Anomaly-Influence Debiasing (AD) Teacher, guided by the alignment between topological and gradient-based interaction matrices.
Extensive experiments on four benchmark datasets demonstrate that HealSplit consistently outperforms ten state-of-the-art defenses, achieving superior robustness and defense effectiveness across diverse attack scenarios.
\end{abstract}

%\begin{links}
%    \link{Code}{https://aaai.org/example/code}
%    \link{Datasets}{https://aaai.org/example/datasets}
%    \link{Extended version}{https://aaai.org/example/extended-version}
%\end{links}

\section{Introduction}

Split Federated Learning (SFL) \cite{thapa2022splitfed} integrates the strengths of Federated Learning (FL) \cite{liu2024vertical,yazdinejad2024robust} and Split Learning (SL) \cite{vepakomma2018split,lin2024efficient}, offering enhanced privacy protection and reduced computational overhead.  An SFL architecture comprises client-side and server-side models. Each client performs the forward pass locally using its client-side model, and then transmits resulting smashed data (intermediate representations) along with corresponding labels to the server. The server-side model performs the rest of the forward and backward computations, producing gradients concerning the smashed data. These gradients are then returned to clients to update their local models. During this process, client-side updates are aggregated by the Fed server, while server-side updates are aggregated by the main server.
%
%Split Federated Learning (SFL) combines the advantages of Federated Learning (FL)\cite{liu2024vertical,yazdinejad2024robust} and Split Learning (SL)~\cite{vepakomma2018split,lin2024efficient}, providing enhanced privacy and reduced computational overhead. In SFL, each client executes the initial forward pass using its client-side model and sends the resulting smashed data (intermediate representations) along with labels to the server. The server-side model completes the remaining forward and backward computations and returns gradients to the client, which are used to update the local model. Client-side updates are aggregated by the Fed server, while server-side updates are aggregated by the main server.

Although SFL is recognized as a robust and privacy-preserving learning paradigm \cite{chen2022fedmsplit}, recent studies have revealed its susceptibility to various data poisoning attacks \cite{fang2020local,wu2024evaluating}. These attacks aim to compromise the learning process by manipulating sophisticated malicious data or modifying model weights, ultimately degrading the performance of the global model or causing misclassifications. The collaborative and split nature of SFL introduces multiple potential attack surfaces \cite{ma2022shieldfl}, including local features \cite{tolpegin2020data}, labels \cite{gajbhiye2022data,ismail2023analyzing}, smashed data \cite{wu2024evaluating}, and client-side model weights \cite{fang2020local,khan2022security}, each of which can be exploited to disrupt training or corrupt model integrity.

To defend against these sophisticated data poisoning attacks in SFL, existing defense strategies remain inadequate, primarily because they are adapted from traditional FL settings. Techniques such as Krum and Multi-Krum (MKRum) \cite{blanchard2017machine}, Trimmed Mean and Median \cite{yin2018byzantine}, and Bulyan \cite{guerraoui2018hidden} primarily rely on statistical aggregation to filter out anomalous or malicious gradients. More advanced defenses, including FLTrust \cite{cao2020fltrust}, DnC \cite{shejwalkar2021manipulating}, FedDMC \cite{mu2024feddmc}, and ShieldFL \cite{ma2022shieldfl}, enhance robustness through trust-based scoring, dimensionality reduction, structural modeling, or encrypted similarity matching. However, these methods typically assume access to full model updates or raw gradients from all clients \cite{yazdinejad2024robust}. Such an assumption does not hold in the SFL setting due to its architectural split and the transmission of smashed data. Consequently, their defensive effectiveness is significantly compromised in SFL. Furthermore, most defenses are designed to address isolated attack vectors, limiting their generalizability against the diverse and complex threat landscape inherent in SFL \cite{ma2022shieldfl}.

\begin{figure*}[t]
	\centering
	\includegraphics[width=0.9\textwidth,height=6.6cm]{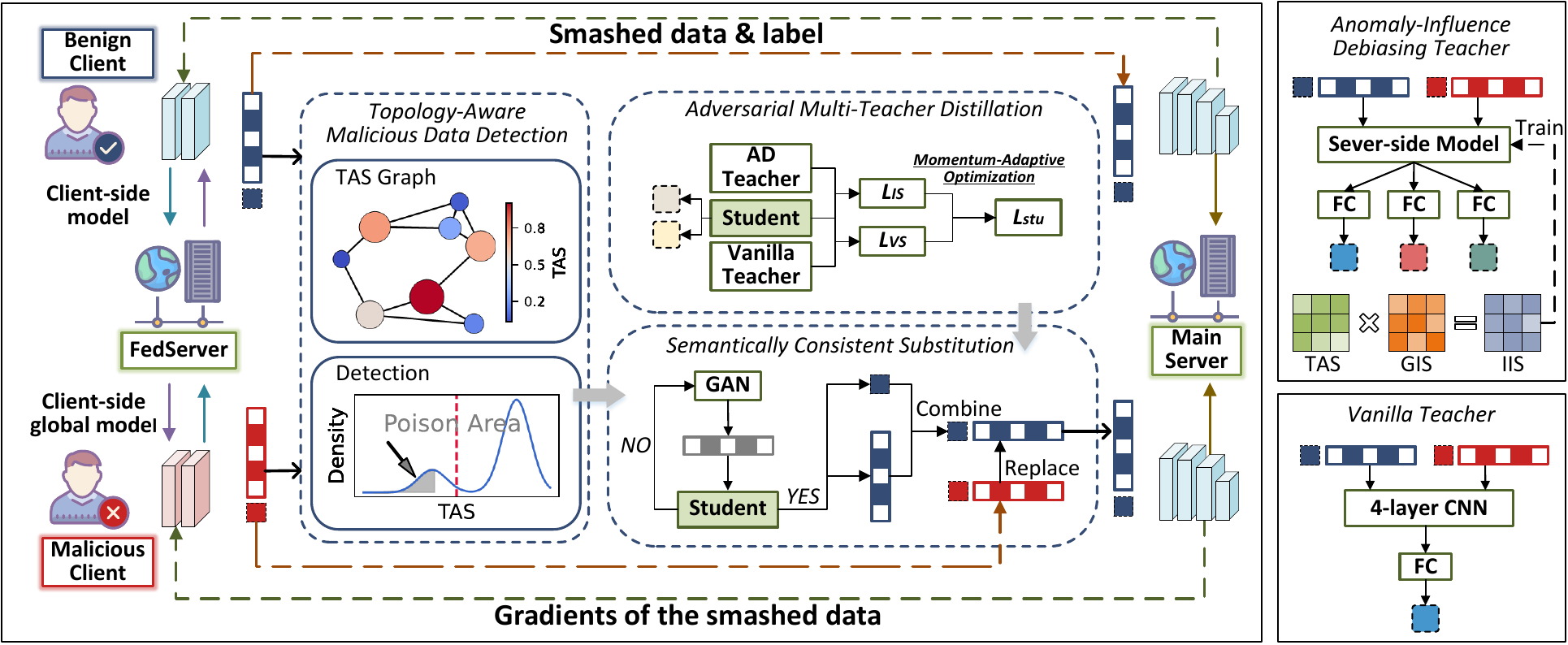}
	\vspace{-2mm}
	\caption{The framework of HealSplit. HealSplit first detects poisoned samples using a KNN-based TAS, and then employs a GAN to generate substitute smashed data. These substitutes are subsequently validated by a consistency validation student model, which is trained via adversarial multi-teacher distillation to ensure semantically consistent substitution.}
	\vspace{-3mm}
	\label{fig:HealSplit}
\end{figure*}

In this paper, we propose HealSplit, a unified defense framework that delivers end-to-end protection against diverse and sophisticated data poisoning attacks in SFL by seamlessly integrating detection and recovery mechanisms. In the architecture of SFL, the smashed data transmitted from clients to server serves as the primary medium for poisoning attacks across various stages of the learning process~\cite{wu2024evaluating}. Accordingly, HealSplit focuses on securing the smashed data to defend against a broad spectrum of poisoning attacks. As shown in Fig.~\ref{fig:HealSplit}, to effectively detect poisoned samples that exhibit anomalous connectivity patterns in the smashed data, a topology-aware detection module is proposed. This module computes a topological anomaly score (TAS) using Personalized PageRank (PPR)~\cite{gasteiger2018predict} on a $k$-nearest neighbors (KNN) graph.

To further rectify the deviations induced by data poisoning attacks, HealSplit incorporates a GAN-based module~\cite{huang2024gan} to generate high-quality substitute representations for detected poisoned samples. These synthetic representations are then validated for semantic consistency using a dedicated student model. The student model is trained via adversarial multi-teacher distillation~\cite{sunmulti}, with supervision provided by two complementary sources:
(1) the Anomaly-Influence Debiasing (AD) Teacher, which regulates inter-task information propagation through an inter-task influence matrix designed to mitigate anomaly-induced bias. This matrix combines the TAS with Gradient Interaction Scores (GIS), enabling the model to selectively propagate information along reliable and structurally consistent task-label paths; and (2) the Vanilla Teacher, which preserves semantic integrity by modeling the distribution of clean data. Contributions from both teachers are dynamically balanced using a momentum-adaptive optimization strategy. Additionally, we theoretically prove that HealSplit reduces gradient variance on the server side by improving gradient similarity.

To comprehensively evaluate the robustness and generalization of HealSplit, we conduct extensive experiments across multi-vector attacks (e.g., DP + SP), heterogeneous data distributions (i.e., IID~$\leftrightarrow$~non-IID), model architectures (e.g., ResNet18~$\leftrightarrow$~VGG16), and adaptive attack strategies. The results show that HealSplit consistently outperforms state-of-the-art defenses, which often fail under dynamic and challenging real-world conditions. Even in the presence of adaptive attacks, it maintains quite a high accuracy, showcasing strong resilience and broad applicability in SFL.

Our contributions are summarized as follows:
\begin{itemize}
	\item We propose the first unified defense framework for SFL that effectively tackles five challenging and diverse attack types: label poisoning, data poisoning, smashed data poisoning, weight poisoning, and  multi-vector poisoning.
	
	\item We introduce a topology-aware detection mechanism that constructs a graph over smashed data and computes a TAS, enabling the detection of poisoned samples by capturing both local and global structural anomalies.
	
	\item We propose a consistency validation student to verify GAN-generated replacements and ensure semantic fidelity. It is optimized via a momentum-adaptive strategy under an adversarial distillation framework.

	\item Extensive experiments on four benchmark datasets demonstrate that HealSplit consistently outperforms ten classical and advanced baselines in both defense efficacy and robustness.
\end{itemize}

\section{Related Work}
\paragraph{Defenses of Split Federated Learning.}
Classic defenses adopt statistical aggregation to mitigate malicious gradients, including Krum and Mkrum \cite{blanchard2017machine}, Trim-Mean and Median \cite{yin2018byzantine}, and Bulyan \cite{guerraoui2018hidden}. In contrast, advanced defenses emphasize robust aggregation and anomaly detection. FLTrust \cite{cao2020fltrust} assigns trust scores based on the alignment between client updates and the server model, and normalizes update magnitudes to reduce the impact of malicious inputs. DnC \cite{shejwalkar2021manipulating} identifies outliers via principal component analysis and leverages dimensionality reduction for efficiency. Feddmc \cite{mu2024feddmc} detects malicious clients through dimensionality reduction, noise-resistant clustering, and self-ensemble correction. ShieldFL \cite{ma2022shieldfl} incorporates two-trapdoor homomorphic encryption for secure aggregation and employs cosine similarity to detect and suppress encrypted model poisoning.

However, these methods are adaptations of FL defenses for the SFL setting. Recent work \cite{wu2024evaluating} has shown that FL defenses are not consistently effective against poisoning attacks in SFL.

\paragraph{Adversarial Distillation.}
Knowledge distillation \cite{hinton2015distilling} is an effective technique for transferring knowledge from a large teacher model to a smaller student model. Recently, adversarial distillation \cite{goldblum2020adversarially,zhao2024mitigating} has gained traction for not only improving model compression but also enhancing robustness \cite{singh2023revisiting,angarano2024domain} and fairness \cite{chai2022fairness,li2024dual} through adversarial training \cite{ganin2015unsupervised}. Among them, DTDBD \cite{li2024dual}, which mitigates domain bias via a dual-teacher framework that balances an unbiased teacher for de-biasing and a clean teacher for domain knowledge transfer. Similarly, B-MTARD \cite{zhao2024mitigating} refines adversarial training by integrating a clean and a robust teacher, employing entropy-based and normalization loss balancing to enhance both accuracy and robustness.

\section{Background}
\subsection{Problem Statement}

Let the SFL system consist of \( N \) clients \( \{c_i\}_{i=1}^N \), each holding a private dataset \( \mathcal{D}_i = \mathcal{D}_i^{\mathrm{tr}} \cup \mathcal{D}_i^{\mathrm{te}} \sim \mathcal{P}_i \), where \( \mathcal{P}_i \) denotes the local data distribution of client \( c_i \). The training and test sets are \( \mathcal{D}_i^{\mathrm{tr}} = \{(x_j, y_j)\}_{j=1}^{m_i^{\mathrm{tr}}} \) and \( \mathcal{D}_i^{\mathrm{te}} = \{(x_j, y_j)\}_{j=1}^{m_i^{\mathrm{te}}} \), where \( x_j \) and \( y_j \in \mathcal{Y} \) denote input features and labels, and 
$m_i^{\mathrm{tr}}$ and $m_i^{\mathrm{te}}$ denote the number of training and test samples, respectively. Clients \( \mathcal{C} = \{c_i\}_{i=1}^N \) are partitioned into benign \( \mathcal{C}_{\text{ben}} \) and malicious \( \mathcal{C}_{\text{att}} = \mathcal{C} \setminus \mathcal{C}_{\text{ben}} \).

Each client maintains a local model \( g_{\theta_{c_i}} \), which maps inputs to smashed data \( z_j = g_{\theta_{c_i}}(x_j) \sim \mathcal{Q}_i \). These smashed data are sent to a server-side model \( h_{\theta_{s_i}} \) for forward computation. In back-propagation, the resulting gradients are used to update \( \theta_{c_i} \). To preserve privacy, the updated client-side and server-side models are sent to the federated server and the main server for aggregation, respectively. The complete model for client \( c_i \) is \( f_{\theta_i} = h_{\theta_{s_i}} \circ g_{\theta_{c_i}} \).

\textbf{Defender Objective.} The defender's goal is to achieve robustness against diverse poisoning attacks while maintaining clean performance. This objective is formulated as a regularized optimization problem:
\begin{equation}
	\min_{\theta} \mathbb{E}_{(x,y) \sim \mathcal{D}^{te}} \left[ \ell(f_\theta(x), y) \right] + \mu \mathcal{R}_{\text{robust}}(\theta),
\end{equation}
where \( \ell(\cdot, \cdot) \) is a task-specific loss function (e.g., cross-entropy), \( \mathcal{R}_{\text{robust}}(\theta) \) measures the model's sensitivity to poisoning behaviors, and \( \mu > 0 \) controls the trade-off between accuracy and robustness.

\subsection{Threat Model} 
In SFL, attackers with varying levels of knowledge and capability can launch diverse and sophisticated data poisoning attacks targeting different stages of the learning pipeline. We categorize these attacks into five major types:

\textbf{Label Poisoning (LP, $\mathcal{A}_{1}$):} The ground-truth label is perturbed as  
\( y_j' = (y_j + \delta_y) \bmod C \),  
where \( C \) is the number of classes and \( \delta_y \) is the label shift.

\textbf{Data Poisoning (DP, $\mathcal{A}_{2}$):} The local dataset  are modified as   
\( \mathcal{D}_k' = \{({x}_j', y_j)\}_{j=1}^{m'} \),  
where \( {x}_j' = x_j + \delta_x \) and \( \delta_x \) denotes input perturbation.

\textbf{Smashed Poisoning (SP, $\mathcal{A}_{3}$):} The smashed data are modified as  
\( z_j' = g_\phi(x_j) + \delta_z \),  
where  \( g_\phi(x_j) \) represents the smashed data and \( \delta_z \) is the feature-level perturbation.

\textbf{Weight Poisoning (WP, $\mathcal{A}_{4}$):} The model parameters are manipulated before aggregation, expressed as  
\( \theta' = \theta + \Delta_\theta \), where \( \Delta_\theta \) represents the weight perturbation.

\textbf{Multi-Vector Poisoning:}  This is a composite attack strategy that integrates multiple poisoning techniques, which is defined as:  
\( \mathcal{A}_{Multi} = \bigcup_{i=1}^{4} S_i \mathcal{A}_i, \, S\in\{0,1\}^4 \), where \( S_i \) indicates whether the \( i \)-th attack \( \mathcal{A}_i \) is applied.

\section{Methodology}
\subsection{Topology-Aware Malicious Data Detection}\label{Topology-Aware}
\begin{figure}[t]
	\centering 
	% 左侧大图，底部对齐
	\begin{minipage}[c]{0.44\linewidth} % 左侧大图，占据页面的48%宽度
		\centering
		\includegraphics[width=\linewidth, height=4.8cm]{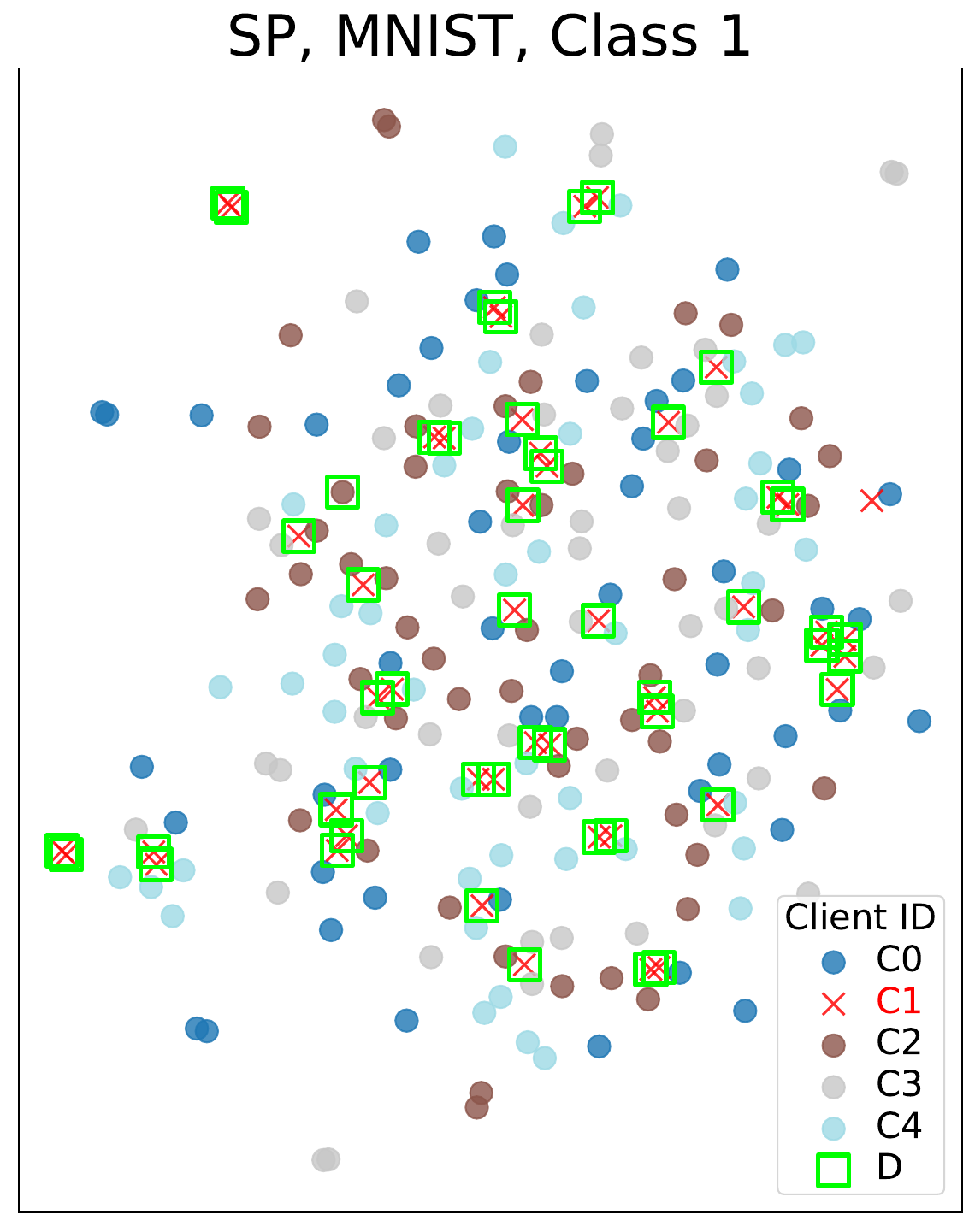} % 这里可以调整图像大小
		\subcaption{Identification of poisoned data.} % 左侧大图的标题
		\label{fig:IdentificationA}
	\end{minipage}
	\begin{minipage}[c]{0.53\linewidth} 
		\centering
		\includegraphics[width=\linewidth, height=2.2cm]{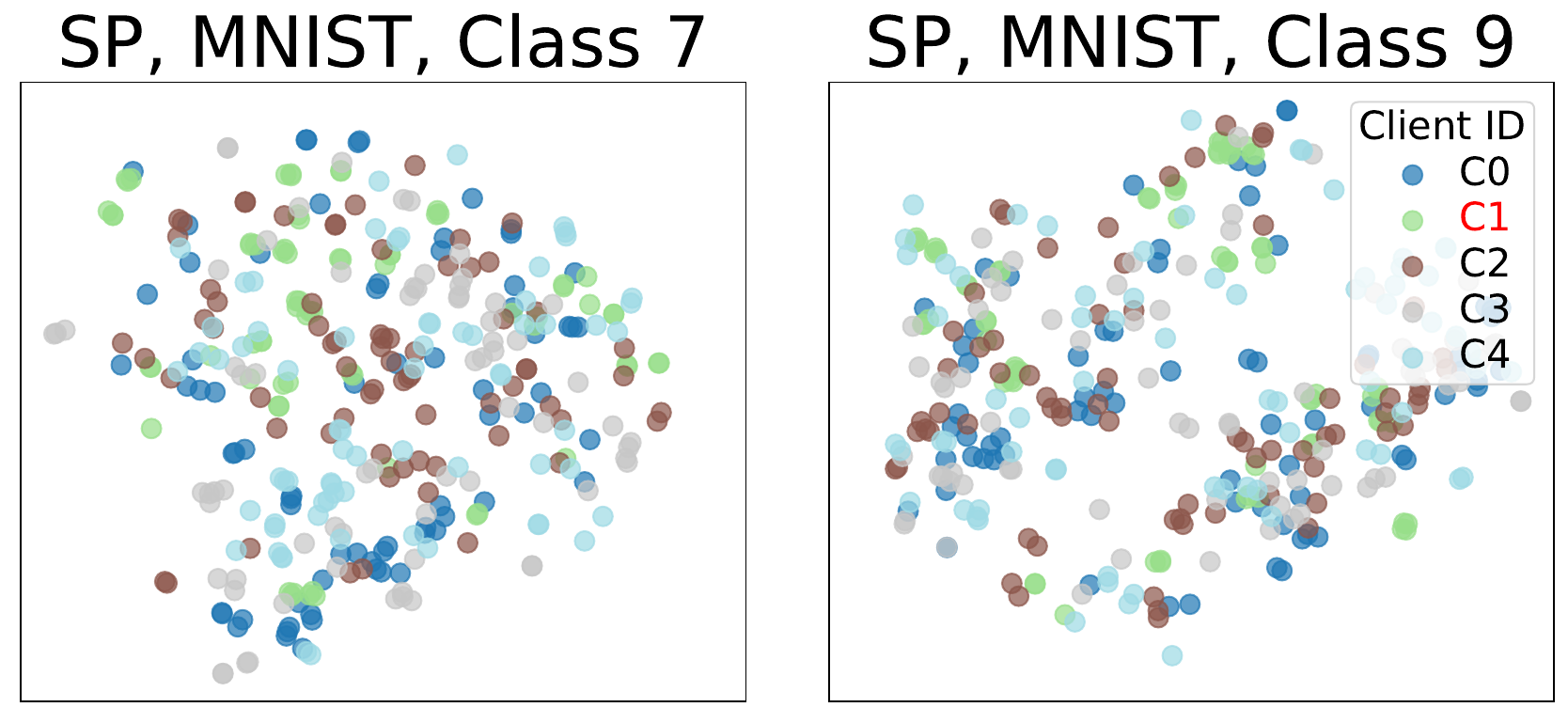}
		\subcaption{Distribution of smashed data.} % 右上图的标题
		\label{fig:IdentificationB}
		\includegraphics[width=\linewidth, height=2.2cm]{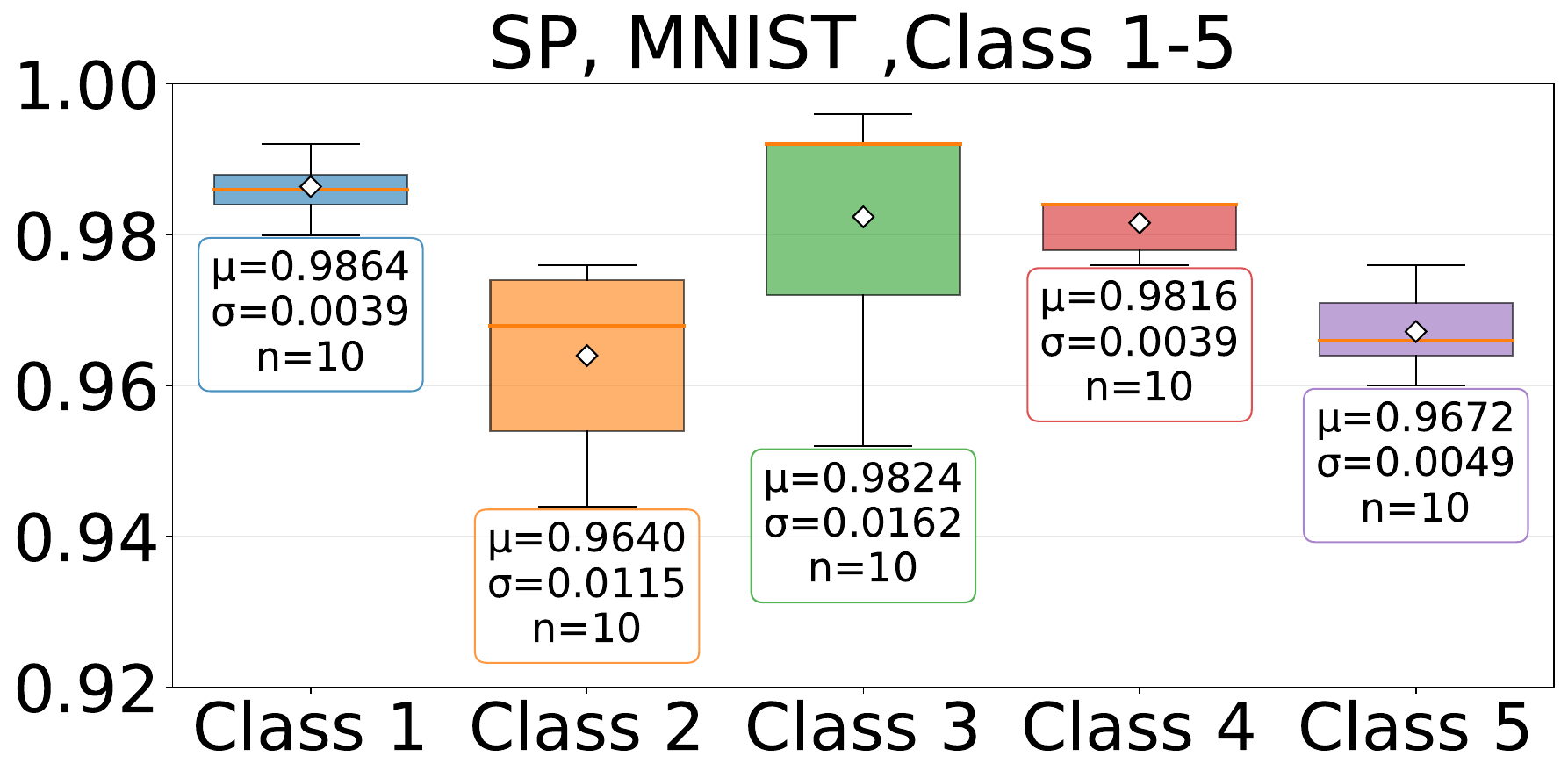}
		\subcaption{Detection rate across classes.} % 右下图的标题
		\label{fig:IdentificationC}
	\end{minipage}
	\vspace{-2mm}
	\caption{Topology-aware malicious data detection: (a) the detection performance, (b) the distribution of smashed data, and (c) detection statistics across different classes. The red crosses denote the smashed data transmitted by malicious client $c_1$, while $D$ denotes the detected malicious samples.}
	\label{IPD}
	\vspace{-4mm}
\end{figure}

Inspired by the graph propagation mechanisms in social networks \cite{zhu2024propagation,cui2024propagation}, our detection framework exploits the topological properties of poisoned data in SFL. As shown in Fig. \ref{fig:IdentificationB}, poisoned samples tend to form locally dense, yet globally isolated clusters in the feature space. This is characterized by (1) high feature similarity within malicious samples, and (2) weak connections to benign data. These observations suggest that topology-based detection can effectively identify poisoning patterns.

\subsubsection{Graph Representation.}\label{GR}
Given smashed data and labels \(\mathcal{D} = \{ (z_k, y_k) \}_{k=1}^K\) obtained from SFL rounds, the weighted graph \(G = (V, E)\) is represented by an adjacency matrix \(\mathbf{W}\):
\begin{equation}
	W_{kj} = 
	\begin{cases} 
		\exp\left(-\gamma \|{z}_k - {z}_j\|^2\right), & \text{if } z_j \in \mathcal{N}_k \text{ and } z_k \in \mathcal{N}_j \\
		0, & \text{otherwise}
	\end{cases}
\end{equation}
where \( \gamma = (2\sigma^2)^{-1} \) and \( \sigma \) is the median between all pairs of points in the KNN graph. \( \mathcal{N}_k \) denotes the KNN set of \( {z}_k \).

\subsubsection{Topological Anomaly Score.}

To identify topologically anomalous nodes exhibiting deviations in propagation patterns, we compute the TAS \( r \) using PPR, capturing both local and global graph structures. The TAS is initialized based on node degrees and updated iteratively at each propagation step \( t \):
\begin{equation} \label{eq4}
	\resizebox{\columnwidth}{!}{$
		r_k^{(t+1)} = \mathbb{I}_{[t = 0]} \cdot \frac{1}{d_k + \epsilon}
		+ \mathbb{I}_{[t \geq 1]} \cdot \left( \alpha \displaystyle \sum_{w \in \mathcal{N}(k)} \frac{r_w^{(t)}}{d_w}
		+ (1 - \alpha) v_k \right),
		$}
\end{equation}
where \( \mathbb{I}[\cdot] \) denotes the indicator function, \( d_w \) is the degree of node \( w \), \( v_k \) is the personalized teleportation vector, \( \epsilon \) is a small constant to prevent division by zero, and \( \alpha \) controls the trade-off between local and global propagation.

\subsubsection{Adaptive Threshold.}\label{AT}

For automatic anomaly detection, we apply kernel density estimation to the TAS:
\begin{equation} \label{eq5}
	\hat{f}(r) = \frac{1}{Kh}\sum_{k=1}^K \mathcal{K}\left(\frac{r-r_k}{h}\right),
\end{equation}
where \( \mathcal{K}(\cdot) \) is the Gaussian kernel function, \( r \) denotes the evaluation point, and \( h \) is the bandwidth. The detection threshold $T$ is defined as:
\begin{equation} \label{eq6}
	T = \min\left(\underset{r}{\operatorname*{argmin}} \hat{f}(r), Q_{\rho}(\{r_k\})\right),
\end{equation}
where \( Q_{\rho}(\{r_k\}) \) denotes the \( \rho \)-percentile of the score set \( \{r_k\} \). Data with scores below the adaptive threshold \( T \) are marked as poisoned. Detection results are shown in Fig.~\ref{IPD}.

\vspace{-1mm}
\subsection{Semantically Consistent Substitution}
%To replace detected malicious smashed data, we train a WGAN-GP using the identified clean smashed data:
%\begin{equation} \label{eq16}
%	\begin{aligned}
	%		\mathcal{L}_{\text{G}} =  \mathbb{E}_{\tilde{\mathbf{z}}}[D(\tilde{\mathbf{z}})] 
	%		- \mathbb{E}_{\mathbf{z}}[D(\mathbf{z})] 
	%		+ \zeta \, \mathbb{E}_{\hat{\mathbf{z}}} \left[ \left( \left\| \nabla_{\hat{\mathbf{z}}} D(\hat{\mathbf{z}}) \right\|_2 - 1 \right)^2 \right]
	%	\end{aligned}
%\end{equation}
%where \( \mathbf{z} \) represents clean smashed features, \( \tilde{\mathbf{z}} \) are generated features from latent codes \( \mathbf{l} \sim \mathcal{N}(0, I) \), and \( D(\cdot) \) is the discriminator. \( \hat{\mathbf{z}} = \xi \mathbf{z} + (1-\xi) \tilde{\mathbf{z}} \) is an interpolation, with \( \xi \sim \mathcal{U}(0, 1) \), and \( \zeta \) is the gradient penalty coefficient.

To replace detected malicious smashed data, we train a vanilla GAN using the identified clean smashed data:
\begin{equation} \label{eq16}
	\begin{aligned}
		\mathcal{L}_{\text{D}} &= - \mathbb{E}_{\mathbf{z}}[\log D(\mathbf{z})] - \mathbb{E}_{\tilde{\mathbf{z}}}[\log (1 - D(\tilde{\mathbf{z}}))] \\
		\mathcal{L}_{\text{G}} &= - \mathbb{E}_{\tilde{\mathbf{z}}}[\log D(\tilde{\mathbf{z}})]
	\end{aligned}
\end{equation}
where \( \mathbf{z} \) represents clean smashed features, \( \tilde{\mathbf{z}} \) are generated features, and \( D(\cdot) \) is the discriminator network.

HealSplit synchronizes with SFL at appropriate intervals, using only the smashed data from the current update round to train the GAN. Although the generator aims to approximate the distribution of clean features, limited training data may lead to semantically inconsistent outputs. To ensure reliability, each generated sample is evaluated by a consistency validation student, and only those with high confidence and label consistency are selected to replace poisoned data.

\subsection{Anomaly-Influence Debiasing Teacher} 
In SFL, malicious clients often employ similar attack patterns \cite{alsaheel2021atlas,luo2022frequency}, leading to biased training that disrupts global aggregation \cite{he2024data,alber2025medical}. To capture these patterns, we utilize TAS and GIS to train an AD teacher model. This model dynamically adjusts label influence by amplifying those that facilitate poisoning detection and attenuating those that obscure it, thereby guiding the learning process toward more effective classification.

\subsubsection{Gradient Interaction Score.}

%In multi-task learning, labels from different tasks influence shared model parameters through gradient propagation, resulting in inter-task interactions.\cite{yu2020gradient,sunmulti} To quantify this effect, we define the GIS between the poisoning detection task and others. Let \( \mathcal{T} = \{a, b, c\} \) represent the tasks, where \( a \) corresponds to poisoning detection, \( b \) to client classification, and \( c \) to category classification. Each task \( t \in \mathcal{T} \) has an associated label set \( \mathcal{Y}_t \). For any task pair \( p \in \{(a, b), (a, c)\} \), the GIS between  individual labels from the label sets  \( \mathcal{Y}_{t_x} \) and \( \mathcal{Y}_{t_y} \) is defined as:
%\begin{equation}\label{eq8}
%	\mathrm{GIS}_p(\mathcal{Y}_{t_x}, \mathcal{Y}_{t_y}) = \sum_{\theta} \cos \left( \nabla_{t_x}(\theta), \nabla_{t_y}(\theta) \right),
%\end{equation}
%where $\cos(\nabla_{t_x}(\theta), \nabla_{t_y}(\theta))$ denotes the cosine similarity between the gradients of the loss functions \(L_{t_x}\) and \(L_{t_y}\) with respect to the shared model parameters \(\Phi_\theta\).  \( \mathrm{GIS}_p(\mathcal{Y}_{t_x}, \mathcal{Y}_{t_y}) > 0 \) indicates aligned gradient directions, suggesting a cooperative relationship that facilitates model optimization. In contrast, \( \mathrm{GIS}_p(\mathcal{Y}_{t_x}, \mathcal{Y}_{t_y}) < 0 \) reflects gradient conflict, implying potential interference that may hinder the optimization process.

In multi-task learning, gradients from different tasks propagate through shared parameters, resulting in inter-task interactions. \cite{yu2020gradient,sunmulti}. Inspired by this, we define the GIS to model two types of inter-task interactions: one between poisoning patterns and client identity, and the other between poisoning patterns and category semantics. 

Let \( \mathcal{T} = \{a, b, c\} \) represent the tasks, where \( a \) corresponds to poisoning identification, \( b \) to client identification, and \( c \) to category classification. Each task \( t \in \mathcal{T} \) has a corresponding label set \( \mathcal{Y}_t \). For a task pair \(p = (t_x, t_y) \in \{(a, b), (a, c)\} \), the GIS between individual labels \( y_{t_x} \in \mathcal{Y}_{t_x} \) and \( y_{t_y} \in \mathcal{Y}_{t_y} \) is represented as a matrix \( \mathbf{G}_p \):
\begin{equation}\label{eq7}
	\resizebox{\columnwidth}{!}{%
		$\mathbf{G}_p = \begin{bmatrix}
			\cos \left( \nabla_{t_x}(y_{t_x}^{(1)}), \nabla_{t_y}(y_{t_y}^{(1)}) \right) & \dots & \cos \left( \nabla_{t_x}(y_{t_x}^{(1)}), \nabla_{t_y}(y_{t_y}^{(|\mathcal{Y}_{t_y}|)}) \right) \\
			\vdots & \ddots & \vdots \\
			\cos \left( \nabla_{t_x}(y_{t_x}^{(|\mathcal{Y}_{t_x}|)}), \nabla_{t_y}(y_{t_y}^{(1)}) \right) & \dots & \cos \left( \nabla_{t_x}(y_{t_x}^{(|\mathcal{Y}_{t_x}|)}), \nabla_{t_y}(y_{t_y}^{(|\mathcal{Y}_{t_y}|)}) \right)
		\end{bmatrix}$},
\end{equation}
Where cosine similarity \( \cos(\cdot, \cdot) \) measures the alignment between task gradients. The magnitude of \( \mathbf{G}_p(\cdot, \cdot) \) reflects the degree of gradient alignment between tasks: larger values denote cooperative interactions conducive to unbiased optimization, while smaller values indicate conflicting objectives that may hinder learning.

%\( \mathbf{G}_p(\cdot, \cdot) > 0 \) indicates aligned gradients between label sets of different tasks, suggesting cooperation that facilitates unbiased optimization. In contrast, \( \mathbf{G}_p(\cdot, \cdot) < 0 \) reflects gradient conflict, signaling opposing interactions that may introduce bias and hinder optimal learning.
%Where \( n_x = |\mathcal{Y}_{t_x}| \) and \( m_y = |\mathcal{Y}_{t_y}| \) are the sizes of the label sets. The cosine similarity \( \cos(\nabla_{t_x}, \nabla_{t_y}) \) quantifies the alignment between the gradients of the loss functions \( L_{t_x} \) and \( L_{t_y} \). \( \mathbf{G}_p(i,j) > 0 \) indicates aligned gradients between label sets of different tasks, suggesting cooperation that facilitates unbiased optimization. In contrast, \( \mathbf{G}_p(i,j) < 0 \) reflects gradient conflict, signaling opposing interactions that may introduce bias and hinder optimal learning.

\subsubsection{Loss Function of AD Teacher.}
%method:将Gp作为先验条件 设计标签感知的转移矩阵进行 TAS和Label的关系融合
To mitigate anomaly-induced bias, we control inter-task interference by designing an inter-task influence score matrix \(\mathbf{M}_p\). Specifically, \(\mathbf{G}_p\) serves as a structural prior to guide the construction of a label-aware transition matrix that integrates task-level and label-level relationships. For each task pair \(p\), \(\mathbf{M}_p\) is computed by combining the TAS matrix \(\mathbf{R}\) with the corresponding GIS matrix \(\mathbf{G}_p\):
\begin{equation}
	\resizebox{\columnwidth}{!}{%
		$\mathbf{M}_p = (1-\beta) \mathbf{E}^\top \left( \mathbf{I}_K - \beta \cdot \left( \mathrm{RowNorm}\left( \mathbf{R} \odot (\mathbf{E}\mathbf{G}_p\mathbf{F}^\top) \right) \right) \right)^{-1} \mathbf{F}$
	}
\end{equation}
where $\beta \in (0,1)$ is the restart probability controlling the range of information propagation, $\mathbf{E} \in \{0,1\}^{K \times |\mathcal{Y}{t_x}|}$ and $\mathbf{F} \in \{0,1\}^{K \times |\mathcal{Y}{t_y}|}$ are the node-to-label mapping matrices for tasks $t_x$ and $t_y$, respectively, $\odot$ denotes the Hadamard product, $\mathbf{I}_K$ is the $K \times K$ identity matrix, and $K$, $|\mathcal{Y}{t_x}|$, and $|\mathcal{Y}{t_y}|$ denote the number of nodes and the sizes of the label sets $\mathcal{Y}{t_x}$ and $\mathcal{Y}_{t_y}$, respectively.

The final loss function of AD Teacher is:
\begin{equation} \label{eq9}
	\begin{split}
		\mathcal{L}_{\mathrm{AD}} = \sum_{k=1}^{K} ( 
		&\mathcal{L}_a(\hat{\mathbf{y}}_k^a, \mathbf{y}_k^a)
		+ \lambda_b [\mathbf{M}_{(a,b)}]_{\mathbf{y}_k^a, \mathbf{y}_k^b} \mathcal{L}_b(\hat{\mathbf{y}}_k^b, \mathbf{y}_k^b)  \\
		& + \lambda_c [\mathbf{M}_{(a,c)}]_{\mathbf{y}_k^a, \mathbf{y}_k^c} \mathcal{L}_c(\hat{\mathbf{y}}_k^c, \mathbf{y}_k^c)),
	\end{split}
\end{equation}
where \( \mathcal{L}_a \), \( \mathcal{L}_b \), and \( \mathcal{L}_c \) denote the loss functions for poisoning identification, client identification, and category classification, respectively, with \( \lambda_b \) and \( \lambda_c \) balancing the latter two tasks. The term \( [\mathbf{M}_p]_{\mathbf{y}_k^a, \mathbf{y}_k^b} \) quantifies the influence between the label sets of tasks $a$ and $b$ for sample $k$, where \( \mathbf{y}_k^a \) and \( \mathbf{y}_k^b \) are their respective labels. Similarly, \( [\mathbf{M}_{(a,c)}]_{\mathbf{y}_k^a, \mathbf{y}_k^c} \) captures the influence between tasks \( a \) and \( c \).

%\subsection{Adversarial Multi-Teacher Distillation}
\subsection{Consistency Validation Student}\label{AMD}
Inspired by adversarial~\cite{sauer2024adversarial,sauer2024fast} and multi-teacher distillation~\cite{wen2024class,ma2024let}, we integrate both to enhance smashed data consistency, improving performance and reducing bias~\cite{li2024dual}. Two teachers capture complementary data aspects in an adversarial setup, while a momentum-adaptive design enables the student to learn more robust and generalized representations.

%\subsubsection{AD Teacher vs. Vanilla Teacher}
\subsubsection{Adversarial Multi-Teacher Distillation.}
The Vanilla Teacher is trained with the identified smashed data, the training loss is:
\begin{equation}\label{eq10}
	\mathcal{L}_{\mathrm{Van}} = \sum_{k=1}^{K}  
	\mathcal{L}_a(\hat{\mathbf{y}}_k^a, \mathbf{y}_k^a).
\end{equation}

The adversarial distillation loss for transferring knowledge from both the Vanilla Teacher and the AD Teacher to the student model is:
\begin{equation} \label{eq11}
	\begin{aligned}
		\mathcal{L}_{\mathrm{VS}} = \tau^{2} \cdot \mathrm{KL} (&\mathrm{LogSoftmax} \left( h_{T_{\mathrm{van}}}(z_i) / \tau \right), \\
		& \left. \mathrm{Softmax} \left( h_{S}(z_i) / \tau \right) \right),
	\end{aligned}
\end{equation}
\begin{equation}\label{eq12}
	\begin{aligned}
		\mathcal{L}_{\mathrm{IS}} = \tau^{2} \cdot \mathrm{KL} (&\mathrm{LogSoftmax} \left( h_{T_{\mathrm{AD}}}(z_i) / \tau\right ),\\
		&  \mathrm{Softmax} \left( h_{S}(z_i) / \tau ) \right),
	\end{aligned}
\end{equation}
where \( h_{T_{\mathrm{van}}} \) and \( h_{T_{\mathrm{AD}}} \) represent the Vanilla Teacher and AD Teacher models, respectively. \( h_{S} \) is the student model, and \( \tau \) is the temperature parameter.

%Both distillation losses have the same structure, aiming to transfer knowledge from the respective teacher models to the student model, improving its generalization and robustness.

The Vanilla and AD Teachers serve complementary roles in adversarial distillation: the Vanilla Teacher captures clean semantics to help the student identify valid features, while the AD Teacher focuses on anomalies to guide deviation detection. This synergy enables the student to integrate semantic clarity with anomaly awareness, ensuring robust evaluation of GAN-generated features for consistency and label alignment.

\subsubsection{Momentum-Adaptive Optimization.}
In order to balance the contributions of the AD Teacher and the Vanilla Teacher and to prevent either from dominating. We design a momentum‐adaptive optimization scheme. The total loss of the consistency validation student is defined as:
\begin{equation} \label{eq13}
	\mathcal{L}_{\text{Stu}}  = \sum_{k=1}^{K} ( 
	\mathcal{L}_a(\hat{\mathbf{y}}_k^a, \mathbf{y}_k^a) 
	+ \lambda_b \mathcal{L}_b(\hat{\mathbf{y}}_k^b, \mathbf{y}_k^b
	) + \mu \mathcal{L}_{\text{VS}} + \eta \mathcal{L}_{\text{IS}}),
\end{equation}
where \(\mu,\eta\) weight the contributions from the Vanilla and AD Teachers, respectively.  We update these weights at each iteration via momentum‐based rules:
\begin{equation}
	\mu_t = m \cdot \mu_{t-1} + (1 - m) \cdot \sigma\left(\kappa \cdot \frac{\mathcal{L}_{\text{VS}} - \mathcal{L}_{\text{IS}}}{\mathcal{L}_{\text{VS}} + \mathcal{L}_{\text{IS}} + \epsilon}\right),
\end{equation}
\begin{equation} \label{eq15}
	\eta_t = m \cdot \eta_{t-1} + (1 - m) \cdot \sigma\left(\kappa \cdot \frac{\mathcal{L}_{\text{IS}} - \mathcal{L}_{\text{VS}}}{\mathcal{L}_{\text{IS}} + \mathcal{L}_{\text{VS}} + \epsilon}\right),
\end{equation}
where \( m \in (0,1] \) is the momentum parameter, \( \kappa \) is a scaling factor, \( \sigma(\cdot) \) is the sigmoid function, and \( \epsilon \) is a small constant. This dynamic adjustment mechanism enables a smooth and gradual balance between the contributions of the two teachers, thereby enhancing model stability and overall performance during training. The momentum term facilitates steady updates to \( \mu \) and \( \eta \), preventing abrupt shifts.

\begin{table}[t]
	\centering
	
	\resizebox{\columnwidth}{!}{
		\begin{tabular}{|l|r|l|r|}
			\hline
			\textbf{Parameter} & \textbf{Value} & \textbf{Parameter} & \textbf{Value} \\
			\hline\hline
			Number of clients & 10 & Learning rate & $1\times10^{-4}$ \\
			\hline
			Malicious client ratio  & 20 & Training epochs & 100 \\
			\hline
			Krum/Trim-mean parameter & 10 & Client participation rate & 100 \\
			\hline
			Attack Method & DP+SP  & Batch size & 64 \\
			\hline
			Sparsefed paprameter & 60 & Local epochs & 1 \\
			\hline
			Bottleneck dimension & 3 & Attack epochs & 100 \\
			\hline
			Coordinate updates per round & 30,000 & Attack learning rate & 0.01 \\
			\hline
			Base Model&Resnet-18&Dataset& MNIST\\	\hline
			AD Teacher&Sever-side model & Vanilla Teacher/Student &4-layer CNN\\	\hline
			%		\hline
		\end{tabular}
	}
	\vspace{-2mm}
	\caption{Experimental parameter settings.}
	\label{tab:parameters}
	\vspace{-4mm}
\end{table}

\begin{table*}[t]
	
	\centering
	\resizebox{\textwidth}{!}{
		\begin{tabular}{lcccccccc}
			\toprule
			\textbf{Defense Method} & \textbf{No Attack} & \textbf{DP} & \textbf{WP} & \textbf{SP} & \textbf{LP} & \textbf{DP+SP} & \textbf{WP+SP} & \textbf{LP+SP} \\
			\midrule	%\rowcolor{gray!20}
			\multicolumn{9}{l}{\underline{\textbf{ {Classic Defenses}}}} \\
			FedAvg         & 96.90±0.01 & 10.12±0.85 & 44.74±19.73 &  \textbf{96.90±0.12} & 79.23±0.73 & 9.19±0.18 & 68.22±0.48 & 64.82±2.42 \\
			Trimmed Mean   & \textbf{97.61±0.58} & 11.12±1.25 & 64.82±10.17 & 94.88±0.45 & 79.71±2.32 & 10.98±0.28 & 70.31±2.25 & 68.12±7.47 \\
			Median         & 93.84±1.36 & 46.42±3.39 & 62.90±9.43  & 68.57±3.18 & 80.44±1.81 & 11.45±2.34 & 46.64±8.29 & 59.86±7.67 \\
			Sparsefed      & 96.65±0.54 & 9.51±1.19  & 13.23±1.14  & 20.06±10.16& 74.50±2.24 & 9.35±0.48  &  {75.24±0.85} & 69.75±2.87 \\
			Krum           & 96.66±1.18 & 76.77±3.71 & 15.91±4.86  & 71.62±0.91 &  \underline{82.95±0.71} & 70.48±1.38 & {76.20±0.53} &  {70.68±3.22} \\
			Bulyan         & 96.85±0.62 & 10.64±1.19 & 21.90±1.97  & \underline{96.82±2.62}& 77.26±2.23 & 10.81±2.41 & 73.00±2.56 & 69.23±1.72 \\
			\midrule
			\multicolumn{9}{l}{\underline{\textbf{{Advanced Defenses}}}}\\
			FLTrust        & 96.52±1.03 & 76.48±1.20 & 48.70±6.11  & 94.42±1.18 & 55.56±3.99 & 73.39±1.44 & 11.33±4.89 & 32.41±4.50 \\
			
			DnC            & 97.27±1.00 &  {80.58±1.32} &  {82.18±2.58}  & 95.33±0.98 & 80.43±1.86 & \underline{76.34±2.33} & \underline{78.82±5.65} & \underline{75.33±2.69} \\
			Feddmc      & 92.48±2.39 & 75.80±2.17 & 31.96±4.18  & 90.79±1.65 & 62.53±4.51 &  75.23±1.50 & 30.42±5.80 & 11.51±0.84 \\
			ShieldFL &\underline{97.58±0.84}&
			\underline{83.73±0.90}&	\underline{84.24±1.91}&	96.35±2.56&	78.18±2.57& 75.54±2.39&	75.16±4.30&	12.97±2.17\\
			\textbf{HealSplit} & 97.17±1.27 & \textbf{96.86±0.77} & \textbf{95.99±1.69} & 96.75±0.86 & \textbf{96.72±0.64} & \textbf{93.88±0.60} & \textbf{92.44±0.73} & \textbf{93.88±1.38} \\
			\bottomrule
			
	\end{tabular}}
	\vspace{-2mm}
	\caption{Results of each defense method under different types of attack.}
	\label{tab:defense_results}
	\vspace{-3mm}
\end{table*}

\vspace{-1mm}
\subsection{Theoretical Foundations of HealSplit}
In SFL, where models are divided into client-side models $g_{\theta_{c_i}}$ and server-side models $h_{\theta_{s_i}}$, we extend the concept of Inter-client Gradient Variance (CGV) \cite{kairouz2021advances,karimireddy2020scaffold} to introduce Inter-Server Gradient Variance (SGV) and establish its upper bound \cite{kairouz2021advances,woodworth2020minibatch}:
\begin{definition}\label{SGV}  
	\textbf{Inter-Server Gradient Variance (SGV)}:  $\text{SGV}(F, \theta_s) = \mathbb{E}_{(z,y) \sim \mathcal{Q}_n} \left\| \nabla_{\theta_s} f_{n}(\theta_s; z, y) - \nabla_{\theta_s} F(\theta_s) \right\|^2.$ SGV is assumed to be upper-bounded, i.e., there exists a constant $\sigma$ such that  $\text{SGV}(F, \theta_s) \leq \sigma^2.$
\end{definition}  
%HealSplit guides the generation and selection of substitute smashed data through adversarial multi-teacher distillation, which evaluates feature quality by filtering out semantically inconsistent and low-confidence samples. Only reliable substitutes aligned with the global distribution $\mathcal{P}$ are used to replace anomalous smashed data $z'$ identified via topological detection. This process not only enhances data quality and generalization but also mitigates SGV.

HealSplit enhances smashed data quality by replacing anomalous smashed data identified through topological detection with reliable substitutes that align with the global distribution \( \mathcal{Q} \). This process not only improves generalization but also effectively reduces SGV. 

To formalize this effect, consider a training round \( T \) where a client \( c_n \) transmits smashed data composed of \( m_n \) clean samples and \( \widehat{m}_n \) poisoned samples drawn from its local training set \( \mathcal{D}_n^{\text{tr}} \). Across all clients, the total number of clean and poisoned samples satisfies \( \sum_{n=1}^N m_n = M \) and \( \sum_{n=1}^N \widehat{m}_n = \widehat{M} \), respectively. Suppose a fraction \( \alpha \) of the clients are malicious. Under the SGV framework, after performing semantically consistent substitution via HealSplit, the following theorem establishes how the objective function constrains gradient dissimilarity:% (proof in Appendix~C):

\begin{theorem}  
	Under the \textbf{SGV} framework (Definition \ref{SGV}), if the ratio of clean samples in a client's dataset satisfies:
	$	\frac{m_n}{m_n + \widehat{m}_n} = \frac{M}{M + \widehat{M}},$
	then the robust objective $\widehat{F}(\theta_s)$ bounds the gradient dissimilarity:
	$ \text{SGV}(\widehat{F}, \theta_s) = \frac{\alpha^2 M^2}{(\widehat{M}+M)^2}  \left\| \nabla_{\theta_s} f_{n}(\theta_s; z, y) - \nabla_{\theta_s} F(\theta_s) \right\|^2 \leq \text{SGV}({F}, \theta_s).$
\end{theorem}

%\vspace{-2mm}

\section{Experiment}
%In this section, we conduct experiments with the aim of answering the following research questions(RQ):
%
%\textbf{RQ1:} What is the effectiveness of HealSplit in defending against different types of attacks?
%
%\textbf{RQ2:} How does HealSplit’s defense performance vary across different datasets?
%
%\textbf{RQ3:} How do the number of clients and the proportion of malicious clients impact HealSplit’s defense effectiveness?
%
%\textbf{RQ4:} How does HealSplit perform on non-IID datasets compared to IID datasets?
%
%\textbf{RQ5:} How does the size of the base model affect HealSplit’s defense performance?

\begin{figure}[!t]
	\centering
	\begin{minipage}{0.23\textwidth}
		\centering
		\includegraphics[width=\textwidth]{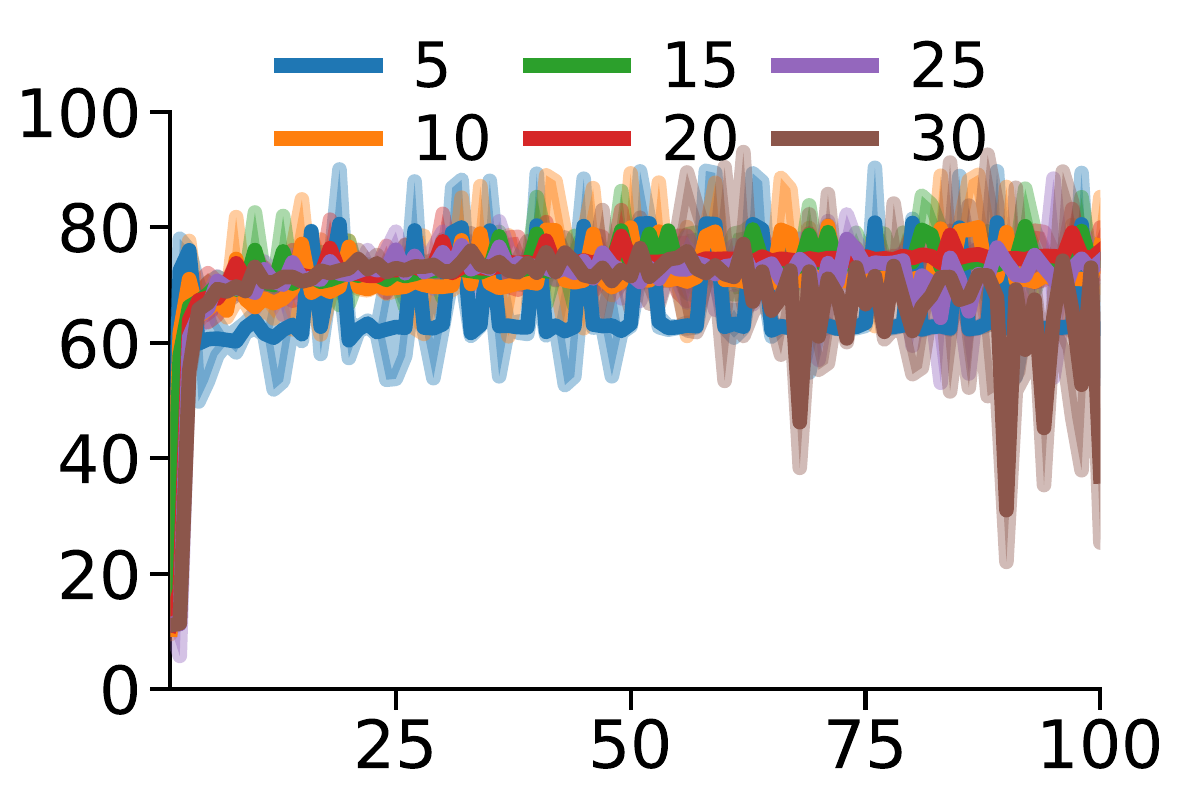}
		\captionsetup{labelformat=empty}
		\vspace{-1.5mm}
		(a) DnC
	\end{minipage}%
	\hfill
	\begin{minipage}{0.23\textwidth}
		\centering
		\includegraphics[width=\textwidth]{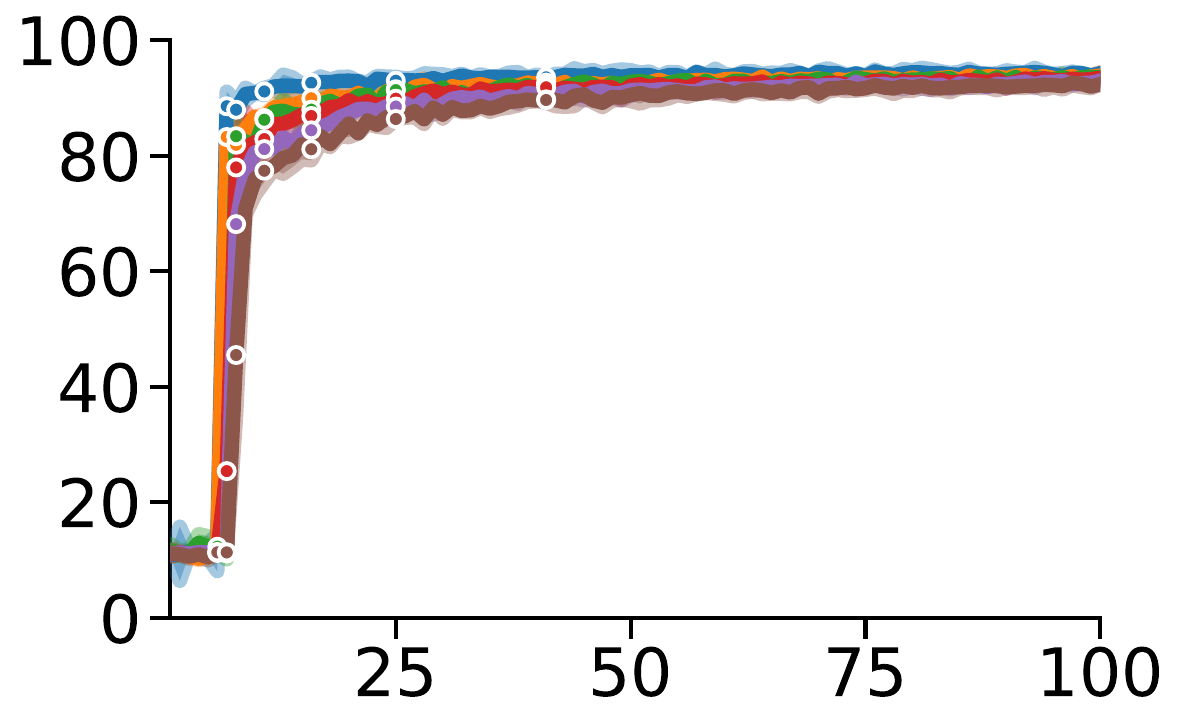}
		\captionsetup{labelformat=empty}
		\vspace{-5mm}
		(b) HealSplit
	\end{minipage}
	%\vspace{-2mm}
	\caption{Defense efficacy across varying client numbers. The circles represent the number of update rounds for the defense model.}
	\label{fig:client_numbers}
	\vspace{-2mm}
\end{figure}

\begin{figure}[!t]
	\centering
	\begin{minipage}{0.23\textwidth}
		\centering
		\includegraphics[width=\textwidth]{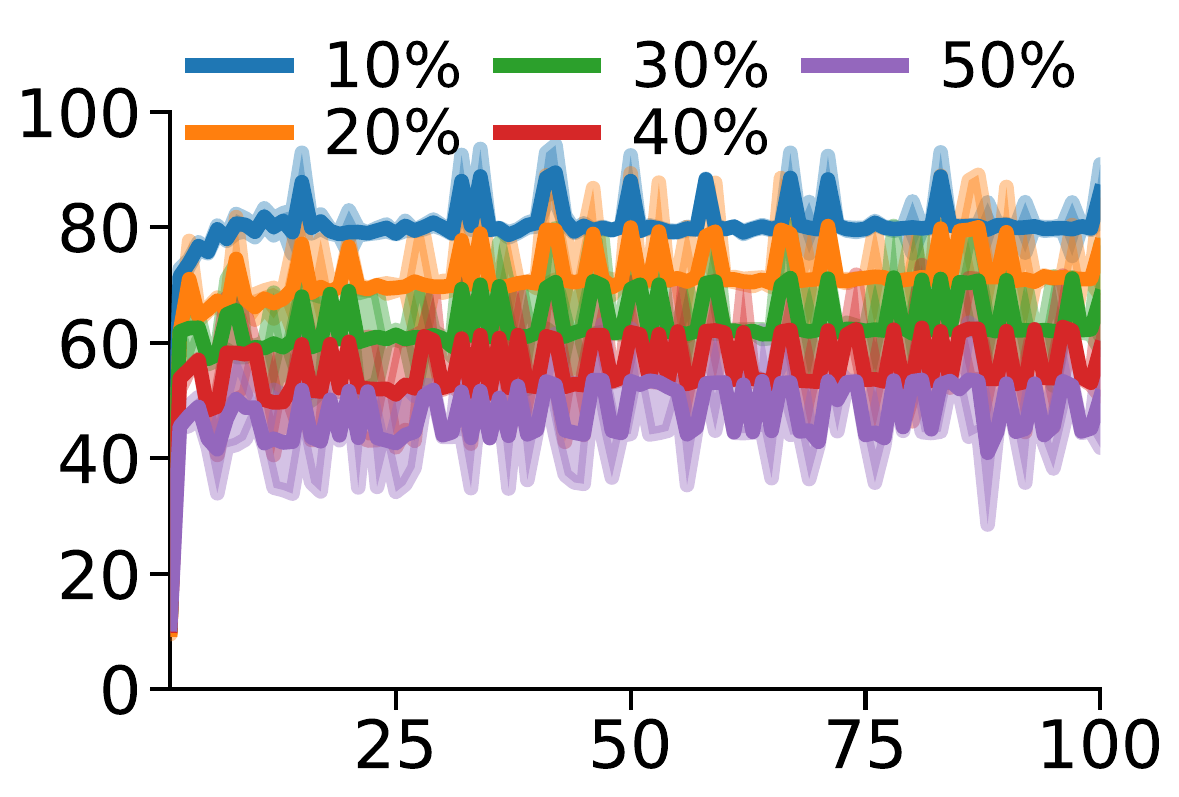}
		\captionsetup{labelformat=empty}
		\vspace{-1.5mm}
		(a) DnC
	\end{minipage}%
	\hfill
	\begin{minipage}{0.23\textwidth}
		\centering
		\includegraphics[width=\textwidth]{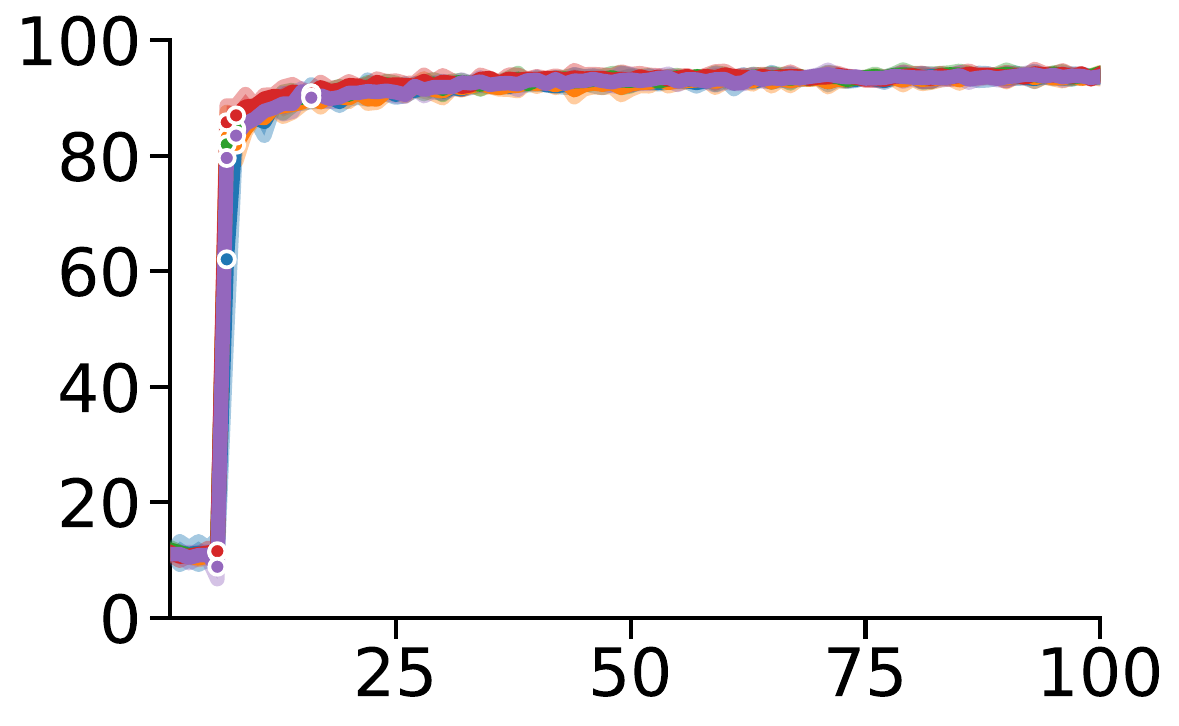}
		\captionsetup{labelformat=empty}
		\vspace{-5.5mm}
		(b) HealSplit
	\end{minipage}
	\caption{Defense efficacy across varying proportions of malicious clients.}
	\label{fig:malicious_proportions}
	\vspace{-4mm}
\end{figure}

\begin{figure}[t]
	\centering
	\includegraphics[width=0.48\textwidth]{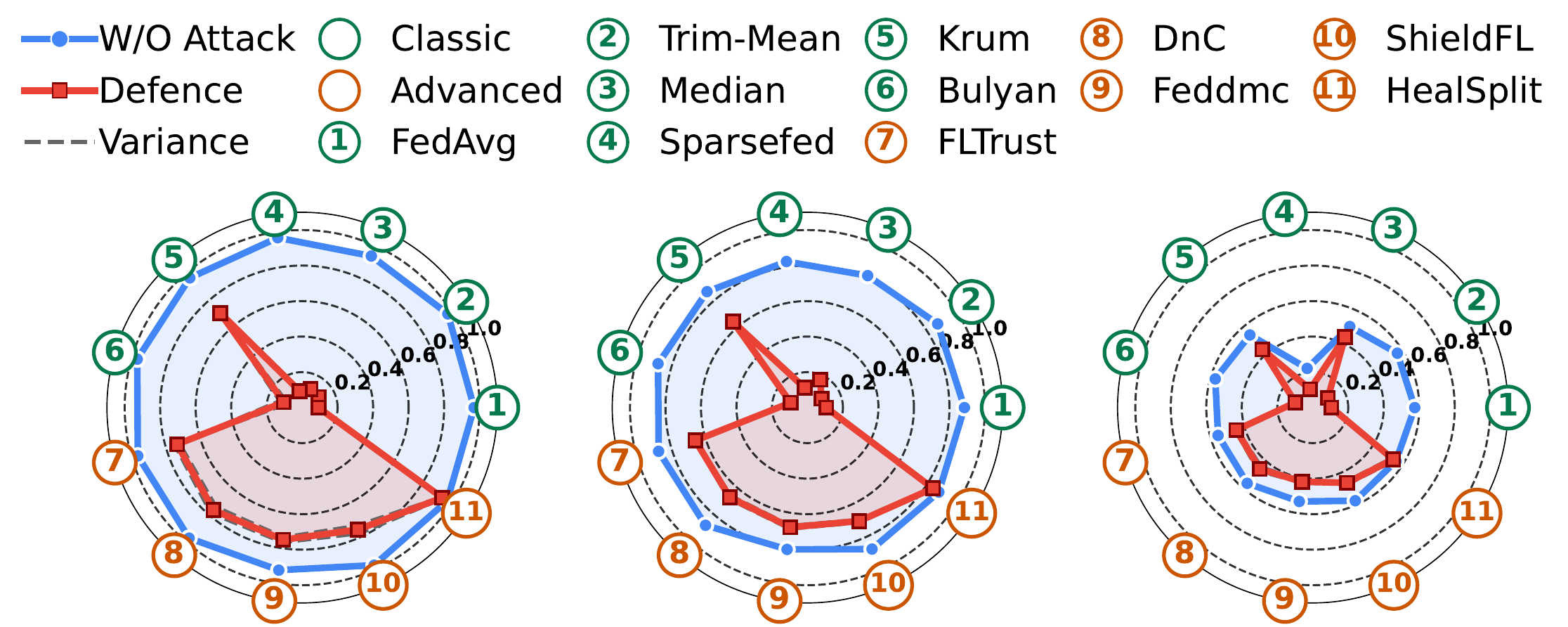}
	
	\vspace{1mm}  % 可选：调整图和标注的垂直距离
	\hspace*{-12mm} 
	\begin{minipage}[t]{0.32\textwidth}
		\centering
		(a) MNIST
	\end{minipage}%
	\hspace*{-30mm} 
	\begin{minipage}[t]{0.32\textwidth}
		\centering
		(b) F-MNIST
	\end{minipage}%
	\hspace*{-30mm} 
	\begin{minipage}[t]{0.32\textwidth}
		\centering
		(c) CIFAR-10
	\end{minipage}
	\vspace{-4mm}
	\caption{Defense efficacy across datasets.}
	\label{fig:Across_Datasets}
	\vspace{-4mm}
\end{figure}
\subsection{Experimental Setup}
\subsubsection{Baselines and Metrics.} 
We assess four attack strategies (DP, WP, SP, and LP) and their combinations (DP + SP, WP + SP, and LP + SP). For comparison, we evaluate ten defense methods: 
FedAvg \cite{mcmahan2017communication}, Trim-Mean, Median, Sparsefed \cite{panda2022sparsefed}, Krum, Bulyan, FLTrust, DnC, Feddmc \cite{mu2024feddmc}, and ShieldFL. 
%FedAvg \cite{mcmahan2017communication}, Trim-Mean \cite{yin2018byzantine}, Median \cite{yin2018byzantine}, Sparsefed \cite{panda2022sparsefed}, Krum \cite{blanchard2017machine}, Bulyan \cite{guerraoui2018hidden}, FLTrust \cite{cao2020fltrust}, DnC \cite{shejwalkar2021manipulating}, Feddmc \cite{mu2024feddmc}, and ShieldFL\cite{ma2022shieldfl}. 
The primary evaluation metric is the reduction in accuracy across all task test sets, which reflects the effectiveness of each defense. %Detailed information is provided in Appendix D.1.

\subsubsection{Datasets and Models.} We evaluate HealSplit on four image datasets: MNIST~\cite{lecun1998mnist}, F-MNIST~\cite{xiao2017fashion}, CIFAR-10~\cite{krizhevsky2009learning}, and HAM10000~\cite{tschandl2018ham10000}. Among these, HAM10000 follows a non-IID distribution, while the others are IID. To further assess HealSplit's robustness under non-IID conditions, we construct a Non-IID version of MNIST, denoted as M-$q$. The degree of data heterogeneity among clients in M-$q$ is controlled by a parameter $q$, with larger $q$ indicating greater non-IIDness. We experiment with three commonly used architectures: ResNet-18 (R18)~\cite{he2016deep}, ResNet-152 (R152), and VGG16~\cite{simonyan2014very}. %Detailed dataset construction and statistical analysis are provided in Appendix D.2 and D.3.

\subsubsection{SFL System Settings.} Unless otherwise specified, each dataset is partitioned among 10 clients, with 20\% acting as malicious participants. The SFL framework is trained for 100 epochs using FedAvg as the default aggregation strategy, under a combined DP and SP attack scheme. Default system configurations are listed in Table~\ref{tab:parameters}.% with further implementation details available in Appendix D.4.

%\begin{figure*}[!t]
%	\centering
%	% 左图组
%	\begin{minipage}[t]{0.48\textwidth}
	%		\centering
	%		\begin{minipage}{0.48\textwidth}
		%			\centering
		%			\includegraphics[width=\textwidth,height=2.2cm]{client_number_Dnc.pdf}
		%			\captionsetup{labelformat=empty}
		%			(a) DnC
		%		\end{minipage}%
	%		\begin{minipage}{0.48\textwidth}
		%			\centering
		%			\includegraphics[width=\textwidth,height=2.2cm]{client_number_HS.pdf}
		%			\captionsetup{labelformat=empty}
		%			(b) HealSplit
		%		\end{minipage}
	%		\caption{Defense efficacy across varying client numbers. The circles represent the number of update rounds for the defense model.}
	%		\label{fig:client_numbers}
	%	\end{minipage}
%	\hfill
%	% 右图组
%	\begin{minipage}[t]{0.48\textwidth}
	%		\centering
	%		\begin{minipage}{0.48\textwidth}
		%			\centering
		%			\includegraphics[width=\textwidth,height=2.2cm]{ratio_DnC.pdf}
		%			\captionsetup{labelformat=empty}
		%			(a) DnC
		%		\end{minipage}%
	%		\begin{minipage}{0.48\textwidth}
		%			\centering
		%			\includegraphics[width=\textwidth,height=2.2cm]{ratio_HS.pdf}
		%			\captionsetup{labelformat=empty}
		%			(b) HealSplit
		%		\end{minipage}
	%		\caption{Defense efficacy across varying proportions of malicious clients.}
	%		\label{fig:malicious_proportions}
	%	\end{minipage}
%\end{figure*}

\subsection{Experimental Results}

\subsubsection{Robustness under Diverse Threats.}
Our first set of experiments evaluates HealSplit's robustness against varying attack strategies, as illustrated in Table \ref{tab:defense_results}.

HealSplit demonstrates consistently strong defense performance, maintaining over 92\% accuracy across all attack scenarios with minimal degradation. It significantly outperforms advanced baselines, which often exhibit large accuracy fluctuations under different attack types. Unlike conventional defenses tailored to specific threats, HealSplit remains robust even against challenging composite attacks.

Notably, state-of-the-art methods such as FLTrust fail under combined attacks like WP+SP, with accuracy dropping to 11.33\%, exposing critical vulnerabilities in the SFL setting. In contrast, HealSplit is inherently attack-agnostic: it requires no prior knowledge of the attack type and avoids fixed defense assumptions.

Moreover, HealSplit detects anomalous client behavior in real time and adaptively adjusts decision thresholds during training, eliminating the need for manual hyperparameter tuning required by other methods such as Krum's neighbor count or SparseFed's norm clipping threshold.

\begin{table}[t]
	\vspace{-2mm}
	\centering
	\resizebox{\columnwidth}{!}{
		\begin{tabular}{lcccc}
			\toprule		
			Component      & \textbf{MNIST} & \textbf{F-MNIST} &\textbf{CIFAR-10}& \textbf{HAM10k} \\ \midrule
			\textbf{HealSplit}   &\textbf{93.88±0.10}& \textbf{84.11±0.47}&\textbf{53.87±1.11}&\textbf{72.27±0.52}\\
			w/o Vanilla Teacher &90.99±1.16 & 80.64±1.49& 51.27±1.35& 69.64±1.51 \\
			w/o AD Teacher      &  87.34±0.16&  75.17±0.39& 46.40±0.60& 63.20±0.45\\ 
			w/o Distillation    &  74.38±3.20& 69.65±3.79& 42.75±2.66& 59.61±2.49 \\ 
			w/o Adversarial    & 92.74±1.68 & 82.59±1.37& 51.55±1.67& 70.40±1.25  \\\bottomrule
	\end{tabular}}
	\vspace{-2mm}
	\caption{Results of ablation study.}
	\vspace{-4mm}
	\label{tab_ablation}
\end{table}

\subsubsection{Ablation Study.}
Our second set of experiments conducts the ablation study under the strong composite attack DP+SP to assess the contribution of each component in HealSplit. The results are presented in Table \ref{tab_ablation}. 

Among all components, the AD Teacher and the distillation mechanism contribute most significantly to HealSplit's overall robustness. Removing the AD Teacher causes a substantial drop in robustness, indicating its role in mitigating model bias through real-time behavioral adjustment. The distillation mechanism is equally critical, as its removal consistently lowers accuracy across tasks, reflecting its effectiveness in integrating multi-task knowledge and enhancing generalization. Excluding the Vanilla Teacher destabilizes training and increases sensitivity to noisy updates, emphasizing its function as a clean semantic reference. Finally, disabling the adversarial mechanism considerably weakens defense performance, highlighting its importance in strengthening resistance to strong attacks.

\subsubsection{Defense Under System Variations.}
Our third set of experiments evaluates the impact of client configurations, including the number of clients and the proportion of adversaries. The results are presented in Fig. \ref{fig:client_numbers} and Fig. \ref{fig:malicious_proportions}. 

HealSplit achieves consistently better performance than the state-of-the-art method DnC after a few rounds of fine-tuning, demonstrating superior robustness across all settings. As the number of clients increases, HealSplit maintains stable, high accuracy, while DnC suffers significant degradation with noticeable variance. Similarly, when the proportion of adversaries rises, HealSplit shows only a mild decline in performance, in contrast to the sharp accuracy drop observed in DnC, highlighting its stronger resilience to adversarial participation.

%and model scales (RQ5) on defense performance.
\subsubsection{Defense Generalization Across Data.}
\begin{figure}[!t]
	\centering
	\begin{minipage}{0.225\textwidth}
		\centering
		\includegraphics[width=\textwidth]{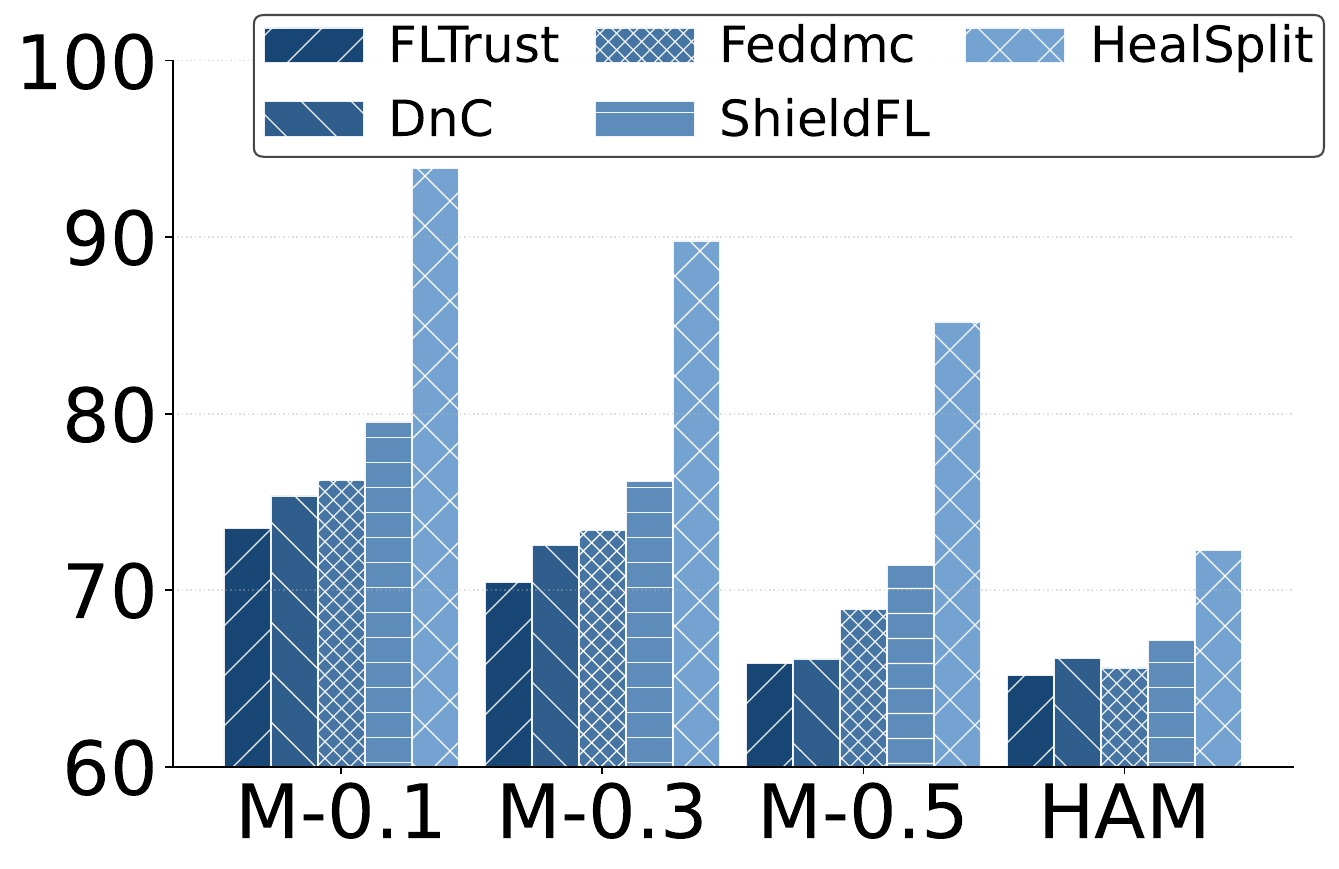}
		\vspace{-6mm}
		\caption{Generalization across non-IID datasets}
		\label{fig:noneiid}
	\end{minipage}%
	%	\hspace{0.05\textwidth}  % 控制两张图片的间隔
	\hspace{2mm}
	\begin{minipage}{0.225\textwidth}
		\centering
		\includegraphics[width=\textwidth]{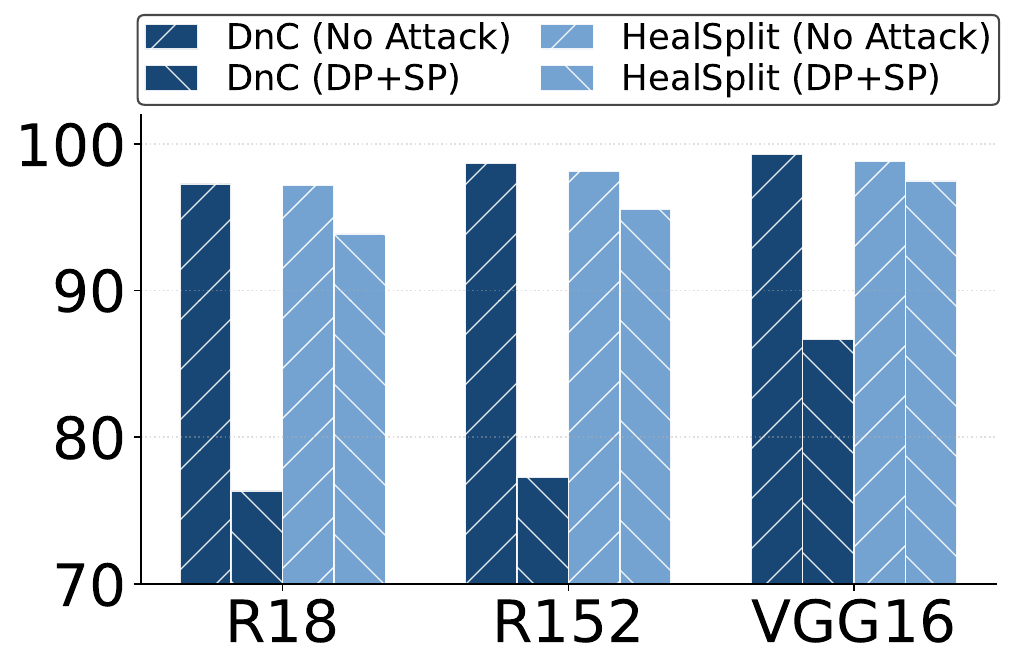}
		\vspace{-6mm}
		\caption{Generalization across models.}
		\label{fig:model_compare}
	\end{minipage}
	\vspace{-3mm}
\end{figure}

Our fourth set of experiments evaluates HealSplit's performance on different datasets and data distributions. The results are presented in Fig. \ref{fig:Across_Datasets} and Fig. \ref{fig:noneiid}.

HealSplit demonstrates strong generalization, consistently outperforming all baselines under both IID and non-IID conditions. On the MNIST dataset, as the data becomes increasingly non-IID, HealSplit maintains stable accuracy above 85\%, while baseline methods suffer significant performance degradation due to increased distribution heterogeneity. On the HAM dataset, which reflects a real-world distributional shift, HealSplit continues to perform robustly, further highlighting its effectiveness across diverse and challenging data environments.

\subsubsection{Defense Generalization Across Model.}
Our fifth set of experiments evaluates HealSplit's generalization ability across different model architectures. The results are presented in Fig. \ref{fig:model_compare}.

Across all tested model architectures, HealSplit consistently outperforms the state-of-the-art baseline DnC, demonstrating strong robustness across varying network structures. Under challenging multi-vector attacks such as DP + SP, it maintains significantly higher accuracy, further underscoring its resilience to architectural variations.

\begin{figure}[t]
	\centering
	\includegraphics[width=0.35\textwidth]{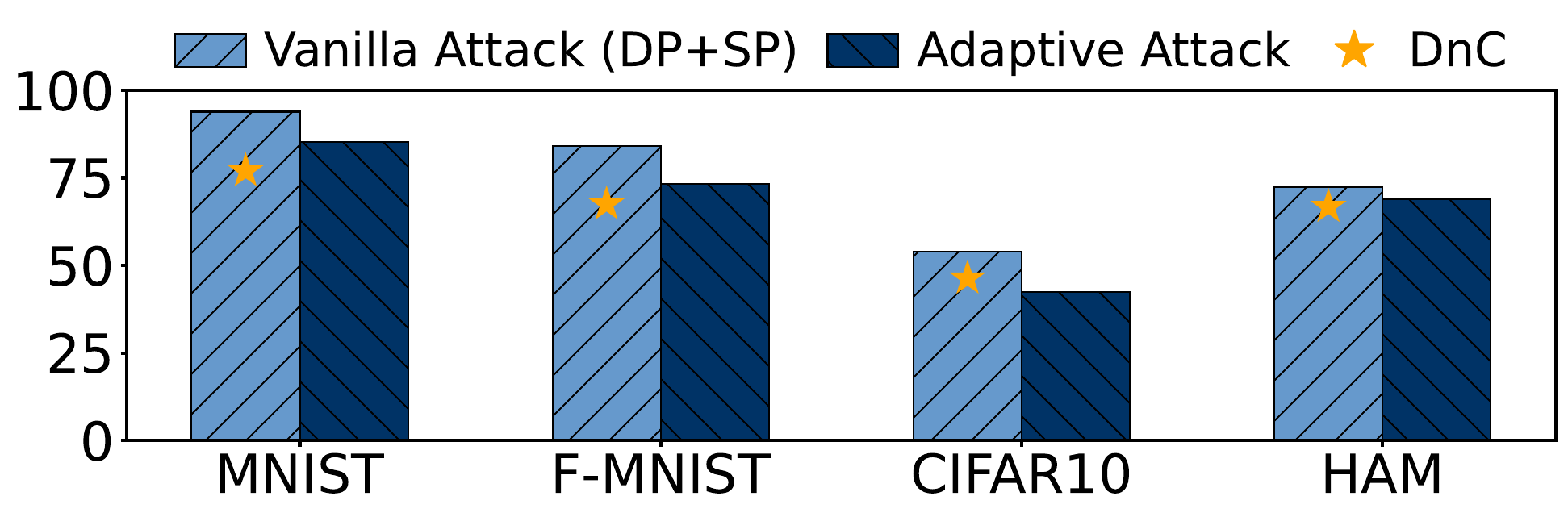}
	\vspace{-2mm}
	\caption{Robustness of HealSplit under adaptive attack.}
	\label{fig:adaptive_attack}
	\vspace{-4mm}
\end{figure}
\subsubsection{Adaptive Attack.} 

Our sixth set of experiments evaluates HealSplit's performance under an adaptive attack to assess its robustness. The results are presented in Fig. \ref{fig:adaptive_attack}.

Specifically, the attacker minimizes the divergence in TAS between poisoned and clean smashed data to evade detection by the topology-aware malicious data detection. These stealthy anomalies subsequently propagate to downstream stages such as GAN training and consistency validation student training, ultimately compromising the overall system.

Although the presence of a stronger and more adaptive threat leads to a noticeable performance drop for HealSplit, the method remains highly competitive. It continues to outperform the strongest existing defenses across multiple datasets, demonstrating superior robustness and generalization even under intensified attack scenarios.

\section{Conclusion}

%\paragraph{Conclusion} We present HealSplit, the first end-to-end defense framework for SFL against data poisoning attacks. By combining topology-aware detection, adversarial multi-teacher distillation, and semantically consistent substitution, HealSplit effectively identifies and replaces poisoned smashed data. Extensive benchmarks demonstrate its superior accuracy and robustness.

%This paper presents HealSplit, the first comprehensive defense framework for diverse and sophisticated data poisoning attacks in SFL. Unlike prior defenses that address isolated attacks or rely on full model access, HealSplit defends against a broad range of poisoning attacks by targeting the smashed data, which serves as the key conduit for adversarial manipulation in SFL. It effectively integrates topology-aware detection, adversarial multi-teacher distillation, and generative recovery into a unified end-to-end system. Extensive experiments demonstrate that HealSplit consistently outperforms existing defenses in robustness and effectiveness, offering a practical and generalizable solution to enhance SFL security.

This paper presents HealSplit, the first comprehensive defense framework against diverse and sophisticated data poisoning attacks in SFL. Unlike prior defenses that address isolated attack types or require full model access, HealSplit mitigates a broad spectrum of poisoning attacks by securing the smashed data, which serves as the primary conduit for adversarial manipulation in SFL. It seamlessly integrates topology-aware detection, adversarial multi-teacher distillation, and generative recovery into a unified end-to-end system. Extensive experiments demonstrate that HealSplit consistently outperforms existing defenses in both robustness and effectiveness, offering a practical and generalizable solution to enhance the security of SFL.

\section{Acknowledgments}
This work was supported by the National Key R\&D Program of China (2023YFA1009500).

\bibliography{aaai2026}

@article{kairouz2021advances,
  title={Advances and open problems in federated learning},
  author={Kairouz, Peter and McMahan, H Brendan and Avent, Brendan and Bellet, Aur{\'e}lien and Bennis, Mehdi and Bhagoji, Arjun Nitin and Bonawitz, Kallista and Charles, Zachary and Cormode, Graham and Cummings, Rachel and others},
  journal={Foundations and trends{\textregistered} in machine learning},
  volume={14},
  number={1--2},
  pages={1--210},
  year={2021},
  publisher={Now Publishers, Inc.}
}

@article{huang2024gan,
  title={The gan is dead; long live the gan! a modern gan baseline},
  author={Huang, Nick and Gokaslan, Aaron and Kuleshov, Volodymyr and Tompkin, James},
  journal={Advances in Neural Information Processing Systems},
  volume={37},
  pages={44177--44215},
  year={2024}
}

@inproceedings{luo2022frequency,
  title={Frequency-driven imperceptible adversarial attack on semantic similarity},
  author={Luo, Cheng and Lin, Qinliang and Xie, Weicheng and Wu, Bizhu and Xie, Jinheng and Shen, Linlin},
  booktitle={Proceedings of the IEEE/CVF conference on computer vision and pattern recognition},
  pages={15315--15324},
  year={2022}
}

@inproceedings{alsaheel2021atlas,
  title={$\{$ATLAS$\}$: A sequence-based learning approach for attack investigation},
  author={Alsaheel, Abdulellah and Nan, Yuhong and Ma, Shiqing and Yu, Le and Walkup, Gregory and Celik, Z Berkay and Zhang, Xiangyu and Xu, Dongyan},
  booktitle={30th USENIX security symposium (USENIX security 21)},
  pages={3005--3022},
  year={2021}
}

@article{simonyan2014very,
  title={Very deep convolutional networks for large-scale image recognition},
  author={Simonyan, Karen and Zisserman, Andrew},
  journal={arXiv preprint arXiv:1409.1556},
  year={2014}
}

@article{mu2024feddmc,
  title={Feddmc: Efficient and robust federated learning via detecting malicious clients},
  author={Mu, Xutong and Cheng, Ke and Shen, Yulong and Li, Xiaoxiao and Chang, Zhao and Zhang, Tao and Ma, Xindi},
  journal={IEEE Transactions on Dependable and Secure Computing},
  year={2024},
  publisher={IEEE}
}

@inproceedings{he2016deep,
  title={Deep residual learning for image recognition},
  author={He, Kaiming and Zhang, Xiangyu and Ren, Shaoqing and Sun, Jian},
  booktitle={Proceedings of the IEEE conference on computer vision and pattern recognition},
  pages={770--778},
  year={2016}
}

@inproceedings{cui2024propagation,
  title={Propagation tree is not deep: Adaptive graph contrastive learning approach for rumor detection},
  author={Cui, Chaoqun and Jia, Caiyan},
  booktitle={Proceedings of the AAAI Conference on artificial intelligence},
  volume={38},
  number={1},
  pages={73--81},
  year={2024}
}

@inproceedings{zhu2024propagation,
  title={Propagation Structure-Aware Graph Transformer for Robust and Interpretable Fake News Detection},
  author={Zhu, Junyou and Gao, Chao and Yin, Ze and Li, Xianghua and Kurths, J{\"u}rgen},
  booktitle={Proceedings of the 30th ACM SIGKDD Conference on Knowledge Discovery and Data Mining},
  pages={4652--4663},
  year={2024}
}

@article{gasteiger2018predict,
  title={Predict then propagate: Graph neural networks meet personalized pagerank},
  author={Gasteiger, Johannes and Bojchevski, Aleksandar and G{\"u}nnemann, Stephan},
  journal={arXiv preprint arXiv:1810.05997},
  year={2018}
}

@article{he2024data,
  title={Data poisoning for in-context learning},
  author={He, Pengfei and Xu, Han and Xing, Yue and Liu, Hui and Yamada, Makoto and Tang, Jiliang},
  journal={arXiv preprint arXiv:2402.02160},
  year={2024}
}

@article{yu2020gradient,
  title={Gradient surgery for multi-task learning},
  author={Yu, Tianhe and Kumar, Saurabh and Gupta, Abhishek and Levine, Sergey and Hausman, Karol and Finn, Chelsea},
  journal={Advances in neural information processing systems},
  volume={33},
  pages={5824--5836},
  year={2020}
}

@article{alber2025medical,
  title={Medical large language models are vulnerable to data-poisoning attacks},
  author={Alber, Daniel Alexander and Yang, Zihao and Alyakin, Anton and Yang, Eunice and Rai, Sumedha and Valliani, Aly A and Zhang, Jeff and Rosenbaum, Gabriel R and Amend-Thomas, Ashley K and Kurland, David B and others},
  journal={Nature Medicine},
  pages={1--9},
  year={2025},
  publisher={Nature Publishing Group US New York}
}

@article{ma2022shieldfl,
  title={ShieldFL: Mitigating model poisoning attacks in privacy-preserving federated learning},
  author={Ma, Zhuoran and Ma, Jianfeng and Miao, Yinbin and Li, Yingjiu and Deng, Robert H},
  journal={IEEE Transactions on Information Forensics and Security},
  volume={17},
  pages={1639--1654},
  year={2022},
  publisher={IEEE}
}

@misc{lecun1998mnist,
  author       = {Yann LeCun and Corinna Cortes and Christopher J. C. Burges},
  title        = {MNIST Handwritten Digit Database},
  year         = {1998},
  note         = {Available: \url{http://yann.lecun.com/exdb/mnist}}
}

@inproceedings{mcmahan2017communication,
  title={Communication-efficient learning of deep networks from decentralized data},
  author={McMahan, Brendan and Moore, Eider and Ramage, Daniel and Hampson, Seth and y Arcas, Blaise Aguera},
  booktitle={Artificial intelligence and statistics},
  pages={1273--1282},
  year={2017},
  organization={PMLR}
}

@inproceedings{panda2022sparsefed,
  title={Sparsefed: Mitigating model poisoning attacks in federated learning with sparsification},
  author={Panda, Ashwinee and Mahloujifar, Saeed and Bhagoji, Arjun Nitin and Chakraborty, Supriyo and Mittal, Prateek},
  booktitle={International Conference on Artificial Intelligence and Statistics},
  pages={7587--7624},
  year={2022},
  organization={PMLR}
}

@article{krizhevsky2009learning,
  title={Learning multiple layers of features from tiny images},
  author={Krizhevsky, Alex and Hinton, Geoffrey and others},
  year={2009},
  publisher={Toronto, ON, Canada}
}

@article{tschandl2018ham10000,
  title={The HAM10000 dataset, a large collection of multi-source dermatoscopic images of common pigmented skin lesions},
  author={Tschandl, Philipp and Rosendahl, Cliff and Kittler, Harald},
  journal={Scientific data},
  volume={5},
  number={1},
  pages={1--9},
  year={2018},
  publisher={Nature Publishing Group}
}

@article{xiao2017fashion,
  title={Fashion-mnist: a novel image dataset for benchmarking machine learning algorithms},
  author={Xiao, Han and Rasul, Kashif and Vollgraf, Roland},
  journal={arXiv preprint arXiv:1708.07747},
  year={2017}
}

@inproceedings{sunmulti,
  title={Multi-Label Node Classification with Label Influence Propagation},
  author={Sun, Yifei and Liu, Zemin and Hooi, Bryan and Yang, Yang and Fathony, Rizal and Chen, Jia and He, Bingsheng},
  booktitle={The Thirteenth International Conference on Learning Representations},
  year = {2025}
}

@inproceedings{karimireddy2020scaffold,
  title={Scaffold: Stochastic controlled averaging for federated learning},
  author={Karimireddy, Sai Praneeth and Kale, Satyen and Mohri, Mehryar and Reddi, Sashank and Stich, Sebastian and Suresh, Ananda Theertha},
  booktitle={International conference on machine learning},
  pages={5132--5143},
  year={2020},
  organization={PMLR}
}

@article{woodworth2020minibatch,
  title={Minibatch vs local sgd for heterogeneous distributed learning},
  author={Woodworth, Blake E and Patel, Kumar Kshitij and Srebro, Nati},
  journal={Advances in Neural Information Processing Systems},
  volume={33},
  pages={6281--6292},
  year={2020}
}

@inproceedings{li2024dual,
  title={Dual-Teacher De-biasing Distillation Framework for Multi-domain Fake News Detection},
  author={Li, Jiayang and Feng, Xuan and Gu, Tianlong and Chang, Liang},
  booktitle={2024 IEEE 40th International Conference on Data Engineering (ICDE)},
  pages={3627--3639},
  year={2024},
  organization={IEEE}
}

@article{singh2023revisiting,
  title={Revisiting adversarial training for imagenet: Architectures, training and generalization across threat models},
  author={Singh, Naman Deep and Croce, Francesco and Hein, Matthias},
  journal={Advances in Neural Information Processing Systems},
  volume={36},
  pages={13931--13955},
  year={2023}
}

@inproceedings{angarano2024domain,
  title={Domain generalization for crop segmentation with standardized ensemble knowledge distillation},
  author={Angarano, Simone and Martini, Mauro and Navone, Alessandro and Chiaberge, Marcello},
  booktitle={Proceedings of the IEEE/CVF Conference on Computer Vision and Pattern Recognition},
  pages={5450--5459},
  year={2024}
}

@article{chai2022fairness,
  title={Fairness without demographics through knowledge distillation},
  author={Chai, Junyi and Jang, Taeuk and Wang, Xiaoqian},
  journal={Advances in Neural Information Processing Systems},
  volume={35},
  pages={19152--19164},
  year={2022}
}

@article{wu2024evaluating,
  title={Evaluating Security and Robustness for Split Federated Learning Against Poisoning Attacks},
  author={Wu, Xiaodong and Yuan, Henry and Li, Xiangman and Ni, Jianbing and Lu, Rongxing},
  journal={IEEE Transactions on Information Forensics and Security},
  year={2024},
  publisher={IEEE}
}

@inproceedings{ganin2015unsupervised,
  title={Unsupervised domain adaptation by backpropagation},
  author={Ganin, Yaroslav and Lempitsky, Victor},
  booktitle={International conference on machine learning},
  pages={1180--1189},
  year={2015},
  organization={PMLR}
}

@article{blanchard2017machine,
  title={Machine learning with adversaries: Byzantine tolerant gradient descent},
  author={Blanchard, Peva and El Mhamdi, El Mahdi and Guerraoui, Rachid and Stainer, Julien},
  journal={Advances in neural information processing systems},
  volume={30},
  year={2017}
}

@inproceedings{yin2018byzantine,
  title={Byzantine-robust distributed learning: Towards optimal statistical rates},
  author={Yin, Dong and Chen, Yudong and Kannan, Ramchandran and Bartlett, Peter},
  booktitle={International conference on machine learning},
  pages={5650--5659},
  year={2018},
  organization={Pmlr}
}

@inproceedings{guerraoui2018hidden,
  title={The hidden vulnerability of distributed learning in byzantium},
  author={Guerraoui, Rachid and Rouault, S{\'e}bastien and others},
  booktitle={International conference on machine learning},
  pages={3521--3530},
  year={2018},
  organization={PMLR}
}

@article{cao2020fltrust,
  title={Fltrust: Byzantine-robust federated learning via trust bootstrapping},
  author={Cao, Xiaoyu and Fang, Minghong and Liu, Jia and Gong, Neil Zhenqiang},
  journal={arXiv preprint arXiv:2012.13995},
  year={2020}
}

@inproceedings{shejwalkar2021manipulating,
  title={Manipulating the byzantine: Optimizing model poisoning attacks and defenses for federated learning},
  author={Shejwalkar, Virat and Houmansadr, Amir},
  booktitle={NDSS},
  year={2021}
}

@inproceedings{gajbhiye2022data,
  title={Data poisoning attack by label flipping on splitfed learning},
  author={Gajbhiye, Saurabh and Singh, Priyanka and Gupta, Shaifu},
  booktitle={International Conference on Recent Trends in Image Processing and Pattern Recognition},
  pages={391--405},
  year={2022},
  organization={Springer}
}

@inproceedings{khan2022security,
  title={Security analysis of splitfed learning},
  author={Khan, Momin Ahmad and Shejwalkar, Virat and Houmansadr, Amir and Anwar, Fatima M},
  booktitle={Proceedings of the 20th ACM Conference on Embedded Networked Sensor Systems},
  pages={987--993},
  year={2022}
}

@article{ismail2023analyzing,
  title={Analyzing the vulnerabilities in splitfed learning: Assessing the robustness against data poisoning attacks},
  author={Ismail, Aysha Thahsin Zahir and Shukla, Raj Mani},
  journal={arXiv preprint arXiv:2307.03197},
  year={2023}
}

@inproceedings{tolpegin2020data,
  title={Data poisoning attacks against federated learning systems},
  author={Tolpegin, Vale and Truex, Stacey and Gursoy, Mehmet Emre and Liu, Ling},
  booktitle={Computer security--ESORICs 2020: 25th European symposium on research in computer security, ESORICs 2020, guildford, UK, September 14--18, 2020, proceedings, part i 25},
  pages={480--501},
  year={2020},
  organization={Springer}
}

@inproceedings{wen2024class,
  title={Class incremental learning with multi-teacher distillation},
  author={Wen, Haitao and Pan, Lili and Dai, Yu and Qiu, Heqian and Wang, Lanxiao and Wu, Qingbo and Li, Hongliang},
  booktitle={Proceedings of the IEEE/CVF Conference on Computer Vision and Pattern Recognition},
  pages={28443--28452},
  year={2024}
}

@inproceedings{ma2024let,
  title={Let all be whitened: Multi-teacher distillation for efficient visual retrieval},
  author={Ma, Zhe and Dong, Jianfeng and Ji, Shouling and Liu, Zhenguang and Zhang, Xuhong and Wang, Zonghui and He, Sifeng and Qian, Feng and Zhang, Xiaobo and Yang, Lei},
  booktitle={Proceedings of the AAAI Conference on Artificial Intelligence},
  volume={38},
  number={5},
  pages={4126--4135},
  year={2024}
}

@inproceedings{sauer2024fast,
  title={Fast high-resolution image synthesis with latent adversarial diffusion distillation},
  author={Sauer, Axel and Boesel, Frederic and Dockhorn, Tim and Blattmann, Andreas and Esser, Patrick and Rombach, Robin},
  booktitle={SIGGRAPH Asia 2024 Conference Papers},
  pages={1--11},
  year={2024}
}

@inproceedings{sauer2024adversarial,
  title={Adversarial diffusion distillation},
  author={Sauer, Axel and Lorenz, Dominik and Blattmann, Andreas and Rombach, Robin},
  booktitle={European Conference on Computer Vision},
  pages={87--103},
  year={2024},
  organization={Springer}
}

@article{lin2024efficient,
  title={Efficient parallel split learning over resource-constrained wireless edge networks},
  author={Lin, Zheng and Zhu, Guangyu and Deng, Yiqin and Chen, Xianhao and Gao, Yue and Huang, Kaibin and Fang, Yuguang},
  journal={IEEE Transactions on Mobile Computing},
  volume={23},
  number={10},
  pages={9224--9239},
  year={2024},
  publisher={IEEE}
}

@article{hinton2015distilling,
  title={Distilling the knowledge in a neural network},
  author={Hinton, Geoffrey and Vinyals, Oriol and Dean, Jeff},
  journal={arXiv preprint arXiv:1503.02531},
  year={2015}
}

@inproceedings{goldblum2020adversarially,
  title={Adversarially robust distillation},
  author={Goldblum, Micah and Fowl, Liam and Feizi, Soheil and Goldstein, Tom},
  booktitle={Proceedings of the AAAI conference on artificial intelligence},
  volume={34},
  number={04},
  pages={3996--4003},
  year={2020}
}

@article{zhao2024mitigating,
  title={Mitigating accuracy-robustness trade-off via balanced multi-teacher adversarial distillation},
  author={Zhao, Shiji and Wang, Xizhe and Wei, Xingxing},
  journal={IEEE Transactions on Pattern Analysis and Machine Intelligence},
  year={2024},
  publisher={IEEE}
}

@article{yazdinejad2024robust,
  title={A robust privacy-preserving federated learning model against model poisoning attacks},
  author={Yazdinejad, Abbas and Dehghantanha, Ali and Karimipour, Hadis and Srivastava, Gautam and Parizi, Reza M},
  journal={IEEE Transactions on Information Forensics and Security},
  year={2024},
  publisher={IEEE}
}

@article{liu2024vertical,
  title={Vertical federated learning: Concepts, advances, and challenges},
  author={Liu, Yang and Kang, Yan and Zou, Tianyuan and Pu, Yanhong and He, Yuanqin and Ye, Xiaozhou and Ouyang, Ye and Zhang, Ya-Qin and Yang, Qiang},
  journal={IEEE Transactions on Knowledge and Data Engineering},
  volume={36},
  number={7},
  pages={3615--3634},
  year={2024},
  publisher={IEEE}
}

@article{vepakomma2018split,
  title={Split learning for health: Distributed deep learning without sharing raw patient data},
  author={Vepakomma, Praneeth and Gupta, Otkrist and Swedish, Tristan and Raskar, Ramesh},
  journal={arXiv preprint arXiv:1812.00564},
  year={2018}
}

@inproceedings{fang2020local,
  title={Local model poisoning attacks to Byzantine-Robust federated learning},
  author={Fang, Minghong and Cao, Xiaoyu and Jia, Jinyuan and Gong, Neil},
  booktitle={29th USENIX security symposium (USENIX Security 20)},
  pages={1605--1622},
  year={2020}
}

@inproceedings{chen2022fedmsplit,
  title={Fedmsplit: Correlation-adaptive federated multi-task learning across multimodal split networks},
  author={Chen, Jiayi and Zhang, Aidong},
  booktitle={Proceedings of the 28th ACM SIGKDD conference on knowledge discovery and data mining},
  pages={87--96},
  year={2022}
}

@inproceedings{thapa2022splitfed,
  title={Splitfed: When federated learning meets split learning},
  author={Thapa, Chandra and Arachchige, Pathum Chamikara Mahawaga and Camtepe, Seyit and Sun, Lichao},
  booktitle={Proceedings of the AAAI conference on artificial intelligence},
  volume={36},
  number={8},
  pages={8485--8493},
  year={2022}
}

\end{document}